\documentclass{article} 
\usepackage[final]{colm2026_conference}

\usepackage{tikz}
\usepackage{amsmath}
\usetikzlibrary{shapes.geometric, arrows.meta, decorations.pathmorphing, positioning, calc, shadows}
\usepackage{microtype}
\usepackage{colortbl}  
\definecolor{divblue}{RGB}{219,234,254}
\usepackage{graphicx}
\usepackage{subcaption}
\usepackage{booktabs} 
\usepackage{stfloats}
\usepackage{hyperref}

\usepackage{amsmath}
\usepackage{amssymb}
\usepackage{mathtools}
\usepackage{amsthm}
\usepackage{enumitem}

\usepackage[capitalize,noabbrev]{cleveref}

\theoremstyle{plain}
\newtheorem{theorem}{Theorem}[section]
\newtheorem{proposition}[theorem]{Proposition}
\newtheorem{lemma}[theorem]{Lemma}
\newtheorem{corollary}[theorem]{Corollary}
\newtheorem{condition}{Condition}
\theoremstyle{definition}
\newtheorem{definition}[theorem]{Definition}

\theoremstyle{remark}
\newtheorem{remark}[theorem]{Remark}

\usepackage[textsize=tiny]{todonotes}

\definecolor{darkblue}{rgb}{0, 0, 0.5}
\hypersetup{colorlinks=true, citecolor=darkblue, linkcolor=darkblue, urlcolor=darkblue}

\newcommand{\bv}{\mathbf{v}}
\newcommand{\bu}{\mathbf{u}}
\newcommand{\bh}{\mathbf{h}}
\newcommand{\bd}{\mathbf{d}}
\newcommand{\bD}{\mathbf{D}}
\newcommand{\bJ}{\mathbf{J}}
\newcommand{\bI}{\mathbf{I}}
\newcommand{\bE}{\mathbf{E}}

\title{Spectral Principal Paths: A Spectral Perspective on Linear Representation Formation in LLMs}

\author{
Bowei Tian$^{1}$ \quad
Xuntao Lyu$^{2}$ \quad
Meng Liu$^{1}$ \quad
Hongyi Wang$^{3}$ \quad
Ang Li$^{1}$
\\
\\
$^1$University of Maryland, College Park \\
$^2$North Carolina State University \\
$^3$Genbio AI
}

\begin{document}


\maketitle

\begin{abstract}
  High-level representations have become a central focus in enhancing AI transparency and control, shifting attention from individual neurons or circuits to structured semantic directions that align with human-interpretable concepts. While the Linear Representation Hypothesis (LRH) suggests that such directions emerge in representations, it remains unclear how these representations originate and why they become increasingly stable across layers. To solve this issue, we introduce the Input-Space Linearity Hypothesis, positing that concept-aligned directions originate in the input space and are steadily maintained with increasing depth. We then propose the Spectral Principal Path (SPP) framework, which formalizes how deep networks progressively distill linear representations along the spectral principal directions. We provide rigorous stability guarantees for the SPP based on the Wedin $\sin\Theta$ perturbation theorem, identifying testable conditions, including spectral gap and context incoherence, that jointly ensure layer-wise directional preservation. By bridging theoretical analysis with empirical evidence, this work identifies a spectral view of how linear representations arise in LLMs, and suggests potential implications for concept-level controllable, robust, and coherent approaches to fairness and transparency in modern AI systems.
\end{abstract}

\section{Introduction}
\label{sec:intro}

Large Language Models (LLMs) have demonstrated strong performance across a wide range of language-related tasks, including question answering \citep{brown2020language,zou2023representation}, code generation \citep{chen2021evaluating}, and reasoning \citep{wei2022chain}. However, the internal mechanisms of neural networks remain opaque. Despite advances in interpretability techniques, how LLMs transform inputs into high-level representations remains poorly understood \citep{nanda2023progress,  weidinger2021ethical}. This opacity raises concerns about reliability in high-stakes domains such as healthcare, finance, and law \citep{bommasani2021opportunities, weidinger2021ethical, conmy2023towards}.
 
Previous works have demonstrated the potential of representations as a new perspective on LLM transparency. 
For example, LLMs have been shown to internally encode linguistic structure across layers, with different layers capturing syntax and semantics at varying levels of abstraction \citep{devlin2019bert, belinkov2022probing}. More recently, structured representations of world knowledge, including spatial and temporal information, have been found to emerge spontaneously in language model hidden states \citep{gurnee2023language, meng2022locating}.
\citet{zou2023representation} further formalized Representation Engineering (RepE), emphasizing its ability to extract meaningful concepts from a model's internal structure and control model behavior. RepE has emerged as a top-down approach to enhance model transparency that focuses on representations rather than individual neurons or circuits, providing a more structured understanding of AI transparency and control. Another important contribution is the Linear Representation Hypothesis (LRH) \citep{park2023linear}: as depth increases, task-relevant concepts become nearly linearly separable in the model's latent space, making them accessible with simple probes or linear edits.

Despite these promising advances, a theoretical understanding of how structured representations emerge and evolve across layers in LLMs remains to be fully established. While approaches such as RepE and LRH have offered valuable insights into the geometry of concept directions, questions remain about why representations become linearly organized and how they propagate through LLMs. A more complete account has yet to emerge --- in particular, it remains unclear whether the linear structure observed in LLM representations is an intrinsic property of the model's learned parameters, or whether it reflects structure already present in the input space prior to any learned transformation.

In this work, we propose Spectral Principal Paths (SPP) as a theoretical framework for investigating the emergence and stability of linear representations in LLMs. We introduce the Input-Space Linearity Hypothesis (ISLH), which suggests that concept-aligned directions originate in the input space and are maintained with increasing depth. These representations propagate through the spectral principal paths, and empirical evidence is in support of this view. This structure offers a potential explanation for why concept directions remain relatively stable and linearly accessible across layers, and sheds light on the theoretical underpinnings of RepE and LRH. We hope this framework contributes a step toward bridging theory and practice in understanding representations in LLMs. Code is released at \href{https://github.com/boweitian/SPP}{https://github.com/boweitian/SPP}.

Our main contributions are as follows:
\begin{itemize}
    \item \textbf{Input-Space Linearity Hypothesis (ISLH).} We extend the Linear Representation Hypothesis beyond the representation space to the input space, showing that concept directions can be traced backward to the input space.

    \item \textbf{Spectral Principal Path (SPP) with rigorous stability guarantees.} We propose the SPP mechanism and provide a formal stability analysis based on the Wedin $\sin\Theta$ perturbation theorem. We identify testable conditions, including spectral gap and context incoherence, under which the SPP direction is provably preserved across layers, with quantitative bounds on angular deviation.

    \item \textbf{Systematic Empirical Validation and Causal Analysis.} We validate the SPP framework across four LLMs, confirming that spectral structures remain robust across structures. Through controlled causal ablations, we provide evidence that both spectral gap and context incoherence are necessary conditions for the emergence and stability of linear concept representations in transformer networks.
\end{itemize}

\section{Related Works}
\label{sec:works}

\subsection{Representation Learning}

Early work on word embeddings shows that neural networks can learn distributed representations that encode semantic relationships and compositional structures \citep{mikolov2013distributed}. Follow-up studies \citep{schramowski2019bert, radford2015unsupervised} further reveal that learned embeddings can implicitly encode abstract dimensions such as commonsense morality, even without explicit supervision. For instance, Radford et al. \citep{radford2015unsupervised} observe that training a language model on product reviews results in the emergence of a sentiment-tracking neuron.

This phenomenon is not unique to language models. \citet{mcgrath2022acquisition} show that similar internal representations can be found in networks trained to play chess. In computer vision, recent studies \citep{caron2021emerging, oquab2023dinov2} demonstrate that both generative and self-supervised training objectives give rise to emergent semantic representations, such as those useful for segmentation tasks, suggesting the emergence of representations is a general property of deep learning systems.

Building on this, \citet{zou2023representation} propose techniques to read and control these internal structures, including Linear Artificial Tomography (LAT) for extracting concept-aligned representations and methods for steering model behavior. Their study shows that RepE-style approaches can be used not only to detect but also to manipulate emergent properties, motivating more systematic efforts to characterize and intervene in high-level model behaviors. Theoretically, \citet{park2023linear} proposes the Linear Representation Hypothesis: task-relevant concepts become nearly linearly separable in the model's latent space.
\vspace{-3mm}

\subsection{Approaches to Interpretability}

Traditional interpretability techniques have focused on methods such as saliency maps \citep{simonyan2013deep, springenberg2014striving, zeiler2014visualizing, zhou2016learning}, feature visualization \citep{szegedy2013intriguing, zeiler2014visualizing}, and mechanistic interpretability \citep{olah2020zoom, olsson2022context, lieberum2023does}. These approaches provide useful local explanations by attributing predictions to specific regions, neurons, or circuits. Saliency maps \citep{simonyan2013deep} identify influential input regions via gradients or activations, but they are often unstable and offer limited insight into how information is distributed across layers. Similarly, feature visualization \citep{szegedy2013intriguing, zeiler2014visualizing} targets individual neurons, which may overlook the broader structure of learned representations.
Mechanistic interpretability \citep{olah2020zoom, olsson2022context, lieberum2023does, zou2023representation} aims to reverse engineer networks into interpretable circuits or algorithmic components. While it has yielded compelling case studies, this approach typically requires substantial manual effort and remains difficult to scale to large models. Moreover, the assumption that complex neural computations can be cleanly decomposed into discrete circuits is theoretically challenging, given the highly distributed and high-dimensional nature of modern representations.

In response to these limitations, more recent work has shifted attention toward the geometry and dynamics of representation spaces. Rather than attributing behavior to individual neurons or circuits, these approaches study how entire representation structures evolve across layers and models. Techniques such as CKA \citep{kornblith2019similarity}, SVCCA \citep{raghu2017svcca}, and Procrustes alignment \citep{gao2021simcse} quantify the similarity and transformation of representations, revealing systematic layer-wise reorganization and shared subspaces across architectures. Complementary lines of research examine how architectural features shape representational flow: residual connections have been shown to support stable feature reuse and smoother optimization trajectories \citep{he2016deep, dehghani2023scaling}. However, these studies largely characterize representational geometry in a descriptive manner, without explaining how such geometric structures are formally propagated, preserved, or amplified across layers.

\section{Preliminaries}
\label{sec:prelim}

\textbf{Linear Representation Hypothesis }\citep{park2023linear}\; We consider a concept $W \in \{0, 1\}$ that has a \emph{linear representation} in a model if there exists a vector $\bar{\gamma}_W$ in the unembedding space $\Gamma$ and a vector $\bar{\lambda}_W$ in the embedding space $\Lambda$ such that for any counterfactual pair $(Y(W=0), Y(W=1))$,
\begin{align}
\gamma(Y(W=1)) - \gamma(Y(W=0)) \in \mathrm{Cone}(\bar{\gamma}_W),
\end{align}
and for any context pair $(\lambda_0, \lambda_1)$ that changes only $W$ and not other causally separable concepts,
\begin{align}
\lambda_1 - \lambda_0 \in \mathrm{Cone}(\bar{\lambda}_W).
\end{align}
where $\mathrm{Cone}(\mathbf{v}) = \left\{ \alpha \mathbf{v} \;:\; \alpha > 0 \right\}$, the embedding space is where input contexts are mapped to high-dimensional vectors before processing, capturing the model's internal representation of the input. The unembedding space is where each output token is represented, and predictions are made by computing inner products between input embeddings and output unembedding vectors. Unless stated otherwise, all discussions pertain to the embedding space $\Lambda$, as our goal is to trace how input linearity propagates through the network.

In the remainder of the paper, we adopt the symmetric encoding $c \in \{-1, 1\}$ for binary concepts, which is equivalent to $c = 2W - 1$.


\begin{figure}[htbp]
\centering
\resizebox{\textwidth}{!}{%
\begin{tikzpicture}[
    >=Stealth,
    concept_arrow/.style={->, line width=3.5pt, draw=orange!85!yellow},
    noise_arrow/.style={->, line width=1.2pt, draw=red!75, decorate, decoration={random steps, segment length=3pt, amplitude=0.8pt, post length=4pt}},
    mean_tube/.style={cylinder, draw=blue!50, fill=blue!5, opacity=0.5, minimum width=4.0cm, minimum height=9.5cm, shape border rotate=0},
    aspect=1.95,
    title_text/.style={align=center, font=\large\bfseries, text=black!80},
    subtitle_text/.style={align=center, font=\small, text=gray!80!black},
    equation_box/.style={draw=blue!40, fill=white, rounded corners, thick, inner sep=8pt, drop shadow={opacity=0.08, shadow xshift=1pt, shadow yshift=-1pt}},
    hypothesis_box/.style={draw=gray!40, fill=gray!5, rounded corners, align=center, inner sep=4pt, text width=2.6cm, font=\footnotesize, drop shadow={opacity=0.05}}
]

\def\YTitle{4.8}
\def\YEq{3.1}
\def\YTube{0}
\def\YHypo{-3.7} 

\node[title_text] at (-3.5, \YTitle) {Phase I: ISLH};
\node[subtitle_text] at (-3.5, \YTitle-0.5) {Input Space Linearity Hypothesis};

\fill[blue!50] (-4.2, 0.7) circle (0.15) (-3.8, 1.0) circle (0.15) (-4.5, 0.9) circle (0.15);
\fill[red!50] (-4.2, -0.7) circle (0.15) (-3.8, -1.0) circle (0.15) (-4.5, -0.9) circle (0.15);
\draw[dashed, gray!70, thick] (-4.2, 0.55) -- (-4.2, -0.55);
\draw[dashed, gray!70, thick] (-3.8, 0.85) -- (-3.8, -0.85);
\draw[dashed, gray!70, thick] (-4.5, 0.75) -- (-4.5, -0.75);

\draw[concept_arrow] (-3.6, \YTube) -- (-0.5, \YTube) node[midway, above=3pt, font=\Large, text=black] {$\mathbf{v}_1^{(0)}$};

\node[title_text] at (4.5, \YTitle) {Phase II: SPP};
\node[subtitle_text] at (4.5, \YTitle-0.5) {Spectral Principal Path};

\node[equation_box] (eq) at (4.5, \YEq) {
    $\mathbf{D}^{(l+1)} \approx \underbrace{\mathbf{D}^{(l)} \bar{\mathbf{J}}^{(l)T}}_{\text{\small Mean Signal (Stable)}} \!\!+\!\! \underbrace{\mathbf{D}^{(l)}_{\text{fluct}}}_{\text{\small Context Noise}}$
};

\node[mean_tube] (tube) at (4.5, \YTube) {};

\node[text=blue!80!black, font=\small\bfseries, fill=white, fill opacity=0.4, text opacity=1, inner sep=3pt, rounded corners] at (4.5, 1.6) {Mean Jacobian Propagation $\bar{\mathbf{J}}^{(l)}$};

\draw[blue!40, dashed, thick] (1.5, -2.0) arc (-90:90:0.4 and 2.0);
\draw[blue!40, dashed, thick] (4.5, -2.0) arc (-90:90:0.4 and 2.0);
\draw[blue!40, dashed, thick] (7.5, -2.0) arc (-90:90:0.4 and 2.0);

\draw[concept_arrow] (-0.5, \YTube) -- (9.5, \YTube) node[pos=0.55, above=3pt, font=\Large, text=black] {$\mathbf{v}_1^{(l)}$};

\draw[noise_arrow] (2.8, 0) -- (3.3, 0.8) node[above=0pt, text=red!80!black, fill=white, fill opacity=0.7, text opacity=1, inner sep=0.5pt, font=\small] {$\mathbf{E}_k^{(l)}$};
\draw[noise_arrow] (4.0, 0) -- (4.6, -0.9);
\draw[noise_arrow] (5.8, 0) -- (6.3, 0.8);
\draw[noise_arrow] (7.0, 0) -- (6.4, -0.9);

\draw[thick, dotted, gray!60] (-0.5, -2.3) -- (9.5, -2.3);
\node[hypothesis_box] (H1) at (2.5, \YHypo) {
    \textbf{Condition I}\\[0.5ex]
    \textbf{Spectral Gap}\\[0.8ex]
    $\sigma_1^{(l)} \gg \sigma_2^{(l)}$\\[0.5ex]
    \textcolor{gray}{Dominant Signal}
};
\draw[->, gray!50, thick] (H1.north) -- (2.5, -2.3);
\node[hypothesis_box] (H3) at (6.5, \YHypo) {
    \textbf{Condition II}\\[0.5ex]
    \textbf{Incoherence}\\[0.8ex]
    $\|\mathbf{\Gamma}^{(l)}\|_{\text{op}} \le \eta^{(l)}$\\[0.5ex]
    \textcolor{gray}{Noise Interferes}
};
\draw[->, gray!50, thick] (H3.north) -- (6.5, -2.3);

\node[title_text] at (12.5, \YTitle) {Phase III: LRH};
\node[subtitle_text] at (12.5, \YTitle-0.5) {Linear Representation Hypothesis};

\draw[concept_arrow] (9.5, \YTube) -- (13.5, \YTube) node[pos=0.6, above=3pt, font=\Large, text=black] {$\mathbf{v}_1^{(L)}$};

\draw[dashed, concept_arrow, opacity=0.5] (9.5, -1.4) -- (13.5, -1.4) node[right, text=black!70] {$\mathbf{v}_1^{(0)}$ \scriptsize (Shifted)};
\draw[<->, thick, darkgray] (12.0, -0.1) -- (12.0, -1.3);

\node[draw=gray!50, fill=white, rounded corners, align=center, font=\footnotesize, drop shadow={opacity=0.05}] at (12.0, -2.5) {
    \textbf{Wedin Theorem Bound}\\
    $\sin \angle(\mathbf{v}_1^{(L)}, \mathbf{v}_1^{(0)}) \le \sum \epsilon_l$
};

\end{tikzpicture}
} 
\caption{Overview of the \textbf{Spectral Principal Path} framework. The conceptual seed $\mathbf{v}_1^{(0)}$ originates in the input space (Phase I: ISLH) and is robustly propagated through network layers via the mean Jacobian $\bar{\mathbf{J}}^{(l)}$ (Phase II). The stability of this path is mathematically guaranteed by (I) Spectral Gap and (II) Context Incoherence, which collectively suppress the destructive influence of context fluctuations. Ultimately, the cumulative angular drift is strictly bounded, culminating in the stable linear representations observed in deep layers (Phase III: LRH).}
\label{fig:overview}
\end{figure}
\section{Formalizing the Emergence of Linear Representations}
\label{sec:formal_proof}

In this section, we provide a formal mathematical framework demonstrating how the Linear Representation Hypothesis (LRH) is naturally derived from the Input-Space Linearity Hypothesis (ISLH) via the formation of highly stable Spectral Principal Paths (SPP). The overview of the Spectral Principal Path framework is shown in Fig.~\ref{fig:overview}. We first define the key objects (\S\ref{sec:islh}--\ref{sec:spp_def}), then state the formal stability guarantee (\S\ref{sec:stability}), and finally derive LRH from ISLH (\S\ref{sec:islh_to_lrh}).

\subsection{Input-Space Linearity Hypothesis (ISLH)}
\label{sec:islh}

Let $X \in \mathbb{R}^{N \times d}$ denote the input sequence of length $N$ with hidden dimension $d$. We focus on the representation of the last token $x_N$, which is typically used for sequence-level predictions.

\begin{definition}[Input-Space Linearity Hypothesis]
\label{def:islh_formal}
Let $c \in \{-1, 1\}$ denote a binary semantic concept (e.g., honest vs.\ dishonest). The ISLH posits that the linear structure described by LRH is a concept-aligned direction already present in the input embedding space:
\begin{equation}
    x_i^{(c)} = \mu_i + c \cdot \alpha_i v_{in} + \epsilon_i, \quad \forall i \in \{1, \dots, N\}
\end{equation}
where $\mu_i \in \mathbb{R}^d$ is the concept-agnostic base semantics of token $i$, $\alpha_i > 0$ is the token-specific expression strength of the concept, $v_{in} \in \mathbb{R}^d$ is the unit concept direction in the input space, and $\epsilon_i \sim \mathcal{N}(0, \sigma^2 I)$ is orthogonal noise. Under this view, the model's role is not to create such directions, but to selectively preserve them across layers via spectral principal paths, which serve as the primary channels through which input-space structure propagates to linear representations.
\end{definition}

\subsection{Information Aggregation via Attention}
\label{sec:layer1}

Initially, the linear concept direction $v_{in}$ is distributed across all $N$ tokens. We first formalize how the attention mechanism in the shallowest layers aggregates this distributed signal into the last token $x_N$.

\begin{lemma}[Attention-Driven Signal Concentration]
\label{lemma:attention_aggregation}
Let $h_1^{(N)} \in \mathbb{R}^d$ be the hidden state of the last token after the first Transformer block. Under the residual attention mechanism, the concept signal is linearly aggregated and amplified:
\begin{equation}
    \Delta h_1^{(N)} = 2 \left( \alpha_N I + W_V \sum_{i=1}^N A_{N,i} \alpha_i \right) v_{in} := \tilde{v}_1
\end{equation}
where $\Delta h_1^{(N)} = \mathbb{E}[h_1^{(N)} | c\!=\!1] - \mathbb{E}[h_1^{(N)} | c\!=\!-1]$ is the concept difference and $A_{N,i}$ are the softmax attention weights.
\end{lemma}

The proof follows by substituting the ISLH formulation into the residual attention update $h_1^{(N)} = x_N + \sum_{i=1}^N A_{N,i} W_V x_i$ (see \Cref{appendix:proof_lemma1}). Thus, the self-attention mechanism acts as a spatial pooling operator, folding the distributed input-space concept $v_{in}$ into a single dominant concentrated direction $\tilde{v}_1$ at the last token position.

\subsection{Spectral Principal Path}
\label{sec:spp_def}

Having established that the concept signal concentrates into a direction $\tilde{v}_1$ at the last token after the first layer, we now formalize how this direction propagates through subsequent layers. The main idea is to study the \emph{population-level} behavior across multiple probing sequences (i.e., paired input sequences designed to elicit contrastive concept expressions), rather than a single input.

\begin{definition}[Difference Matrix and SPP]
\label{def:diff_matrix}
Given $M$ paired sequences $\{(s_k^+, s_k^-)\}_{k=1}^M$ differing only in \emph{concept tokens} (i.e., the tokens whose content determines the concept label $c$), let $\bh_k^{+,(l)}, \bh_k^{-,(l)} \in \mathbb{R}^d$ denote the last-token hidden state at layer $l$. Define the \emph{concept difference vector} $\bd_k^{(l)} = \bh_k^{+,(l)} - \bh_k^{-,(l)}$ and the \emph{difference matrix}:
\begin{equation}
\bD^{(l)} = \begin{pmatrix} (\bd_1^{(l)})^\top \\ \vdots \\ (\bd_M^{(l)})^\top \end{pmatrix} \in \mathbb{R}^{M \times d}.
\end{equation}
The \emph{Spectral Principal Path} (SPP) at layer $l$ is the leading right singular vector $\bv_1^{(l)} \in \mathbb{R}^d$ of $\bD^{(l)}$, corresponding to the largest singular value $\sigma_1^{(l)}$.
\end{definition}

\begin{definition}[SPP Stability]
\label{def:stability}
The SPP is \emph{$\epsilon$-stable} between layers $l$ and $l+1$ if $|\langle \bv_1^{(l)}, \bv_1^{(l+1)} \rangle| \geq 1 - \epsilon$.
\end{definition}

\paragraph{Layer-wise dynamics.} For the $k$-th sequence, the last-token hidden state $\bh_k^{(l)} \in \mathbb{R}^d$ evolves through a nonlinear map whose Jacobian we denote $\bJ_k^{(l)} = \partial \bh_k^{(l+1)} / \partial \bh_k^{(l)} \in \mathbb{R}^{d \times d}$. Due to residual connections, $\bJ_k^{(l)} = \bI + \boldsymbol{\Delta}_k^{(l)}$, where $\boldsymbol{\Delta}_k^{(l)}$ aggregates the attention and MLP gradients. To first order:
\begin{equation}
\label{eq:jacobian_approx}
\bd_k^{(l+1)} \approx \bJ_k^{(l)} \bd_k^{(l)} + \boldsymbol{\xi}_k^{(l)},
\end{equation}
where $\boldsymbol{\xi}_k^{(l)}$ is a higher-order remainder. We decompose the Jacobian perturbation across sequences into a population mean and individual fluctuation:
\begin{equation}
\label{eq:mean_fluct}
\boldsymbol{\Delta}_k^{(l)} = \bar{\boldsymbol{\Delta}}^{(l)} + \bE_k^{(l)}, \quad \text{where } \bar{\boldsymbol{\Delta}}^{(l)} = \frac{1}{M}\sum_{k=1}^M \boldsymbol{\Delta}_k^{(l)}.
\end{equation}
Here $\bE_k^{(l)} \in \mathbb{R}^{d \times d}$ is the context-specific Jacobian fluctuation for the $k$-th sequence, capturing how sequence $k$'s Jacobian deviates from the population mean.

\subsection{Stability of the Spectral Principal Path}
\label{sec:stability}

We now state the formal conditions under which SPP stability is guaranteed. Each assumption is an empirically testable condition rather than a structural consequence of the architecture.

\begin{condition}[Spectral Gap]
\label{ass:gap}
At each layer $l$, the difference matrix $\bD^{(l)}$ has a nontrivial spectral gap: $\sigma_1^{(l)}  \gg \sigma_2^{(l)}$, where $\sigma_1^{(l)}$ and $\sigma_2^{(l)}$ are the largest and second-largest singular values of $\bD^{(l)}$, respectively.
\end{condition}

\begin{condition}[Context Incoherence]
\label{ass:incoherence}
At each layer transition $(l, l{+}1)$, define the per-sample fluctuation
\begin{align}
\mathbf{f}_k^{(l)}
\;=\;
\bigl(\bd_k^{(l+1)} - \bd_k^{(l)}\bigr)
\;-\;
\frac{1}{M}\sum_{j=1}^{M}\bigl(\bd_j^{(l+1)} - \bd_j^{(l)}\bigr),
\end{align}
which captures how the $k$-th sample's layer-to-layer evolution deviates
from the population trend. The empirical second-moment matrix
$\Gamma^{(l)} = \frac{1}{M}\sum_{k=1}^{M}
\mathbf{f}_k^{(l)}(\mathbf{f}_k^{(l)})^\top$
satisfies
$\|\Gamma^{(l)}\|_{\mathrm{op}} \leq \eta^{(l)}$,
where $\eta^{(l)}$ is small relative to $(\sigma_1^{(l)})^2$.
When contexts are diverse, the perturbation directions
$\{\mathbf{f}_k^{(l)}\}$ scatter across the ambient space,
and $\eta^{(l)} = O(\tau^2/d)$
(see \Cref{appendix:incoherence_proof}).
\end{condition}

\begin{theorem}[SPP Stability via Wedin's $\sin\Theta$ Theorem]
\label{thm:spp_stability}
Under Conditions~\ref{ass:gap}--\ref{ass:incoherence}, and assuming the higher-order remainder satisfies $\frac{1}{M}\sum_k \|\boldsymbol{\xi}_k^{(l)}\|^2 \leq \zeta^{(l)} (\sigma_1^{(l)})^2$ with $\zeta^{(l)} \ll 1$, the angular deviation of the SPP between consecutive layers is bounded by:
\begin{equation}
\label{eq:main_bound}
\sin\angle\!\left(\bv_1^{(l+1)}, \bv_1^{(l)}\right) \leq \frac{\sigma_1^{(l)}\bigl(\delta^{(l)} + \sqrt{\eta^{(l)}} + \sqrt{\zeta^{(l)}}\bigr)}{\gamma^{(l+1)}},
\end{equation}
\end{theorem}

where $\delta^{(l)} = \sin \angle(\bar{\bJ}^{(l)} \bv_1^{(l)}, \bv_1^{(l)})$, $\gamma^{(l)} = \sigma_1^{(l)} - \sigma_2^{(l)}$. The proof (given in \Cref{appendix:proof_stability}) proceeds by decomposing $\bD^{(l+1)}$ into a ``signal'' component $\bD^{(l)}\bar{\bJ}^{(l)\top}$ (whose leading direction is close to $\bv_1^{(l)}$) and a ``perturbation'' from context fluctuations and remainder terms, then applying the Wedin $\sin\Theta$ theorem \citep{wedin1972perturbation} to bound the rotation of the leading singular vector.

\begin{corollary}[Global SPP Stability]
\label{cor:global}
If $\sin\angle(\bv_1^{(l+1)}, \bv_1^{(l)}) \leq \epsilon_l$ for each layer, then:
\begin{equation}
\angle\!\left(\bv_1^{(L)}, \bv_1^{(0)}\right) \leq \sum_{l=0}^{L-1} \epsilon_l.
\end{equation}
Under worst-case alignment of per-layer drifts this gives $L\bar{\epsilon}$. However, if the drift directions are approximately independent across layers, the accumulation follows a random walk on the sphere, yielding $O(\sqrt{L}\,\bar{\epsilon})$ (see \Cref{appendix:global_stability}).
\end{corollary}

\begin{remark}[Empirical testability]
\label{rem:testability}
Each condition maps to a direct measurement: \Cref{ass:gap} is verified by computing $\sigma_1^{(l)}/\sigma_2^{(l)}$ at each layer; \Cref{ass:incoherence} is assessed by checking whether the off-principal-direction component of $\bD^{(l+1)} - \bD^{(l)}$ is spectrally diffuse. We report these measurements in \Cref{sec:experiments}.
\end{remark}

\subsection{Deriving the Linear Representation Hypothesis}
\label{sec:islh_to_lrh}

We now bridge ISLH and LRH by tracing the concentrated concept $\tilde{v}_1$ through the entire network.

\begin{theorem}[ISLH to LRH via SPP]
\label{thm:islh_to_lrh}
Given the aggregated input direction $\tilde{v}_1$ (\Cref{lemma:attention_aggregation}) and the stable Spectral Principal Path (\Cref{thm:spp_stability}), the deep representation $h^{(L)}$ satisfies the Linear Representation Hypothesis.
\end{theorem}
\begin{proof}
The total concept perturbation at layer $L$ is:
\begin{equation}
    \Delta h^{(L)} \approx \left( \prod_{l=1}^{L-1} \bJ^{(l)} \right) \tilde{v}_1.
\end{equation}
Projecting onto the SVD of each Jacobian, the accumulated transformation is governed by the cumulative gain along each spectral path:
\begin{equation}
\label{eq:gp}
    \Delta h^{(L)} \approx \sum_{k=1}^d \underbrace{\left( \prod_{l=1}^{L-1} \sigma_k^{(l)} \Theta_k^{(l)} \right)}_{G(\mathcal{P}_k)} \langle v_k^{(1)}, \tilde{v}_1 \rangle \, u_k^{(L-1)},
\end{equation}
where $\Theta_k^{(l)} = |\langle u_k^{(l)}, v_k^{(l+1)} \rangle|$ is the inter-layer alignment factor measuring how well the $k$-th left singular vector at layer $l$ aligns with the $k$-th right singular vector at layer $l{+}1$. By \Cref{thm:spp_stability}, the principal path has $\Theta_1^{(l)} \approx 1$, while spurious paths have $\Theta_k^{(l)} < 1$ for $k > 1$. Combined with the spectral gap $\sigma_1^{(l)} > \sigma_k^{(l)}$, the system acts as a multi-stage power iteration. As $L$ increases, all $k > 1$ terms decay exponentially:
\begin{equation}
    \Delta h^{(L)} \;\propto\; \left( \prod_{l=1}^{L-1} \sigma_1^{(l)} \right) u_1^{(L-1)}.
\end{equation}
The semantic difference collapses into a single 1D subspace $\mathrm{Span}(u_1^{(L-1)})$, satisfying the formal definition of LRH (since $\mathrm{Cone}(\mathbf{v}) \subset \mathrm{Span}(\mathbf{v})$ for any $\mathbf{v}$, and the positive scaling from the singular value products ensures the cone condition holds).
\end{proof}

\section{Experiments}
\label{sec:experiments}
 
We evaluate the SPP framework on four large language models spanning different architectures and scales: \textbf{Qwen 2.5-7B}~(28 layers)~\citep{hui2024qwen2}, \textbf{Llama 3.1-8B-Instruct}~(32 layers)~\citep{grattafiori2024llama}, \textbf{Mistral 7B-Instruct-v0.3}~(32 layers)~\citep{jiang2023mistral7b}, and \textbf{Wizard-Vicuna-30B}~(60 layers)~\citep{wizardvicuna30b2023}.
For each model, we use concept pairs drawn from the COCO dataset (780 pairs; $M{=}80$ samples per concept)~\citep{lin2014microsoft}.
Hidden states are collected at the last token position of each input.
All singular value decompositions are performed on the difference matrix $\mathbf{D}^{(l)} \in \mathbb{R}^{M \times d}$ as defined in Definition~4.3, whose $k$-th row is $\mathbf{d}_k^{(l)} = \mathbf{h}_k^{+,(l)} - \mathbf{h}_k^{-,(l)}$.
 
\subsection{Spectral Gap and Cross-Layer Stability}
\label{sec:exp-spectral-gap}

To verify Condition \ref{ass:gap}, we investigate whether the spectral gap $\rho^{(l)} = \sigma_1^{(l)} / \sigma_2^{(l)}$ of the difference matrix $\mathbf{D}^{(l)}$ is substantially large, and whether its leading singular direction remains structurally stable across layers. For each layer $l$, we compute the spectral gap $\rho^{(l)}$ and compare it against a randomly shuffled baseline. To assess the spatial coherence of the principal path, we evaluate the cross-layer angular stability, defined as the absolute cosine similarity $|\cos(\mathbf{v}_1^{(l)},\, \mathbf{v}_1^{(l+1)})|$ between the leading singular vectors of adjacent layers.  

\begin{figure}[t]
  \centering
  \includegraphics[width=1.01\textwidth]{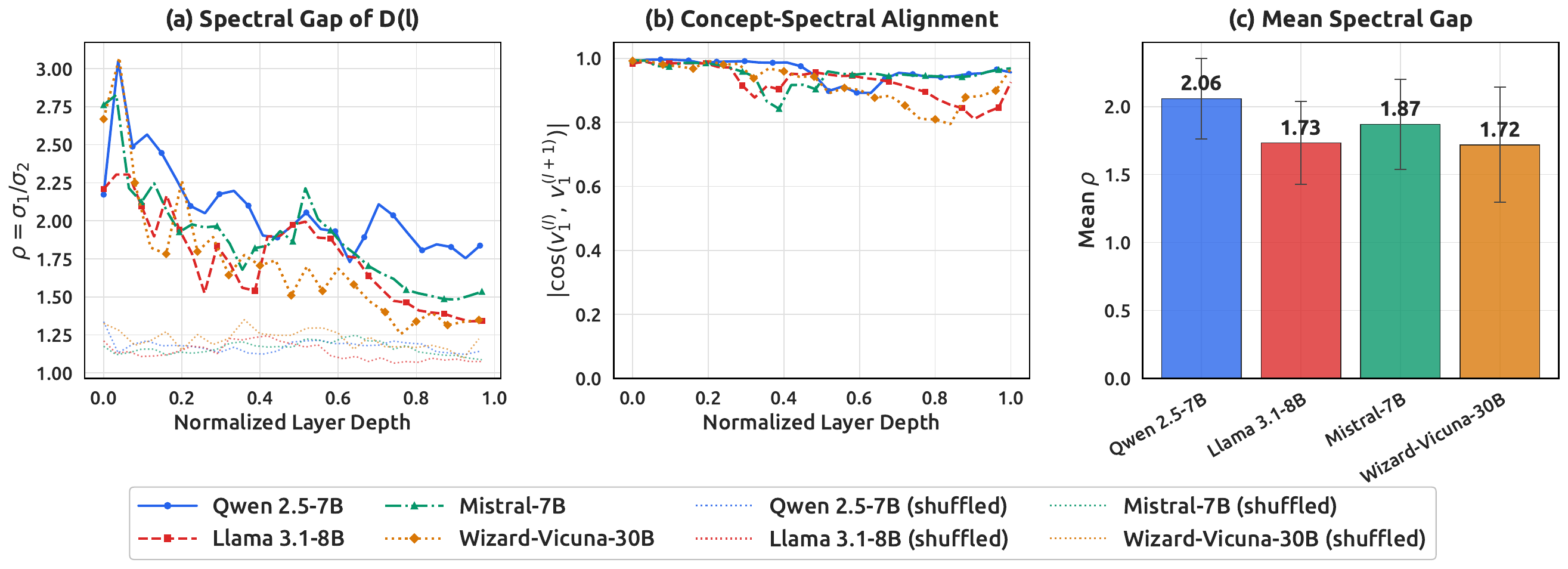}
  \caption{Spectral dominance of the concept direction across four LLMs.
  (a) Spectral gap $\rho^{(l)}$ of the difference matrix per layer, with shuffled baselines (shaded);
  (b) Concept-spectral alignment;
  (c) Mean spectral gap per model.}
  \label{fig:spectral-dominance}
\end{figure}

Our empirical results validate these theoretical requirements across all four models. As shown in Figure~\ref{fig:spectral-dominance}(a) and (c), we observe that the computed spectral gap $\rho^{(l)}$ exhibits a consistent enhancement compared to the randomly shuffled baseline, while this enhancement is especially evident in the first few layers. Furthermore, Figures~\ref{fig:spectral-dominance}(b) demonstrate that the calculated spectral principal path effectively propagates across layers as a highly stable and coherent trajectory, as the majority of layers sees a cosine similarity of over 0.85. Even Wizard-Vicuna-30B, 
with 60 layers, maintains a stable spectral principal path throughout, 
suggesting that the SPP mechanism scales gracefully with network depth. These observations provide direct empirical support for the necessary conditions of SPP formation and further verify the effectiveness of Theorem \ref{thm:spp_stability}.

\subsection{Context Incoherence}
\label{sec:exp-incoherence}

\begin{figure}[t]
  \centering
  \includegraphics[width=1.01\textwidth]{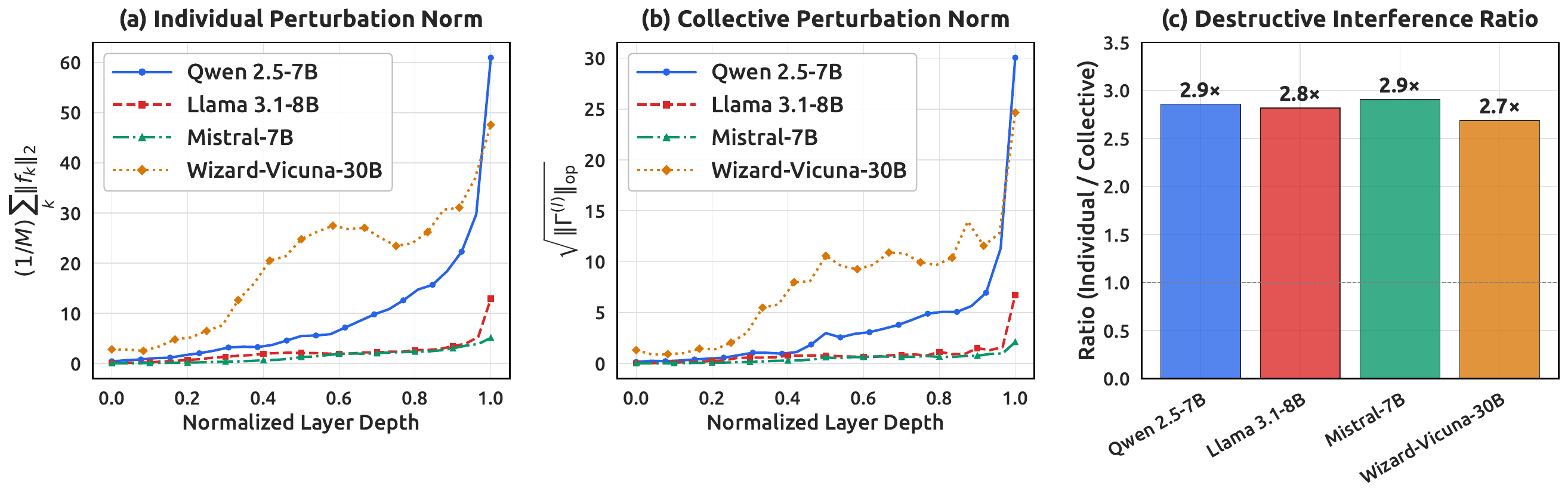}
  \caption{Context incoherence verification.
  (a) Individual perturbation norm. (b) Collective perturbation norm.
  (c) Destructive-interference ratio per model.}
  \label{fig:context-incoherence}
\end{figure}

Next, we verify Condition \ref{ass:incoherence} (Context Incoherence), which posits that individual per-sample perturbations are large but scatter across mutually incoherent directions, rendering their aggregate effect small. To test this in Figure \ref{fig:context-incoherence}, we isolate the sample-specific fluctuation $\mathbf{f}_k^{(l)}$ from the population mean during the layer transition. We then measure the \emph{destructive-interference ratio}, defined as the individual perturbation strength ($\frac{1}{M}\sum_{k=1}^{M} \|\mathbf{f}_k^{(l)}\|_2$) divided by the collective perturbation strength ($\sqrt{\|\mathbf{\Gamma}^{(l)}\|_{\mathrm{op}}}$), where $\mathbf{\Gamma}^{(l)} = \frac{1}{M}\sum_{k=1}^{M} \mathbf{f}_k^{(l)} (\mathbf{f}_k^{(l)})^\top$. We observe that both individual perturbation and collective perturbation increase across depth: 
it is lower in early layers, where perturbations are less isotropically 
distributed, and highest in the later layers. We also observe that while individual perturbation norms are substantially large, the collective perturbation strength remains considerably smaller. Specifically, the destructive-interference ratio is consistently $2.7$--$2.9\times$, exceeding the baseline of $1$ expected for coherent perturbations, indicating substantial but not perfect
incoherence---sufficient for the Wedin bound in
Theorem~\ref{thm:spp_stability} to ensure SPP stability. 

\subsection{Violating Context Incoherence Disrupts SPP Stability}
\label{sec:exp-breaking}

Having established that Condition \ref{ass:incoherence} holds under naturalistic prompting, we further determine whether context incoherence is a \emph{necessary} condition for SPP stability by deliberately violating it. We construct two synthetic evaluation datasets: a \textbf{Diverse} dataset where each sample is drawn from a distinct contextual distribution, ensuring that per-sample residual perturbations are approximately independent across the batch, and a \textbf{Homogeneous} dataset constructed by prepending a shared, nearly identical long prefix to every sample, which forces the per-sample activations to become highly correlated. We evaluate the spectral gap $\rho^{(l)}$ and cross-layer angular stability $|\cos(\mathbf{v}_1^{(l)},\, \mathbf{v}_1^{(l+1)})|$ across layers, where the former measures how cleanly the leading singular direction separates from the residual subspace, and the latter captures the geometric consistency of the principal concept direction across consecutive layers.

As expected, switching from Diverse to Homogeneous contexts consistently degrades both metrics. The mean cross-layer angular stability drops by $0.06$--$0.09$, and the mean spectral gap $\bar\rho$ drops by $0.23$--$0.50$. The mechanism is transparent: under context incoherence, per-sample perturbations approximately cancel out across the batch, preventing any spurious direction from accumulating sufficient variance to challenge the dominance of $\mathbf{v}_1^{(l)}$. Under homogeneous contexts, however, the shared prefix induces correlated perturbations that concentrate spectral energy along a small number of directions, collectively deflecting the leading singular vector away from the true concept signal and simultaneously eroding the spectral gap. This provides causal evidence that context incoherence is structurally necessary for the emergence and stability of linear concept representations along the spectral principal path.
 
\begin{table}[t]
\centering
\caption{SPP metrics under Diverse vs.\ Homogeneous contexts.
$|\cos(v_1^{(l)},\,v_1^{(l+1)})|$: mean cross-layer angular stability (mean $\pm$ std);
$\bar\rho$: mean spectral gap (mean $\pm$ std);
$\Delta$: Homogeneous $-$ Diverse.}
\label{tab:diverse_homo}
\small
\setlength{\tabcolsep}{5pt}
\begin{tabular}{lcccccc}
\toprule
 & \multicolumn{3}{c}{$|\cos(v_1^{(l)},\,v_1^{(l+1)})|$} & \multicolumn{3}{c}{$\bar\rho$} \\
\cmidrule(lr){2-4}\cmidrule(lr){5-7}
Model & Diverse & Homogeneous & $\Delta\!\downarrow$ & Diverse & Homogeneous & $\Delta\!\downarrow$ \\
\midrule
Qwen 2.5-7B
  & $0.853_{\pm.100}$
  & $0.793_{\pm.119}$
  & \cellcolor{divblue}\textbf{-0.060}
  & $1.83_{\pm.48}$
  & $1.33_{\pm.24}$
  & \cellcolor{divblue}\textbf{-0.50} \\
Llama 3.1-8B
  & $0.847_{\pm.139}$
  & $0.763_{\pm.231}$
  & \cellcolor{divblue}\textbf{-0.083}
  & $1.76_{\pm.78}$
  & $1.31_{\pm.10}$
  & \cellcolor{divblue}\textbf{-0.45} \\
Mistral-7B
  & $0.881_{\pm.118}$
  & $0.795_{\pm.195}$
  & \cellcolor{divblue}\textbf{-0.086}
  & $1.67_{\pm.61}$
  & $1.29_{\pm.28}$
  & \cellcolor{divblue}\textbf{-0.37} \\
Wizard-Vicuna-30B
  & $0.785_{\pm.149}$
  & $0.705_{\pm.211}$
  & \cellcolor{divblue}\textbf{-0.080}
  & $1.59_{\pm.46}$
  & $1.36_{\pm.17}$
  & \cellcolor{divblue}\textbf{-0.23} \\
\bottomrule
\end{tabular}
\label{tab:homogenous}
\end{table}


\section{Conclusion}

This work presents a novel spectral perspective that grounds the emergence and
stability of linear representations in LLMs. We introduce the
Input-Space Linearity Hypothesis, which proposes that concept-aligned
directions originate in the input embedding space and are propagated---rather
than created---by the network. Building on this, the Spectral Principal Path framework formalizes how these directions are selectively preserved across
layers through spectrally dominant paths. We provide rigorous
stability guarantees via Wedin's sin\,$\Theta$ Theorem, identifying two
empirically testable conditions---spectral gap and context incoherence---under
which the SPP direction is provably preserved with quantitative angular bounds.
We further formally derive the Linear Representation Hypothesis (LRH) from
ISLH via the SPP, showing that spectral dominance is sufficient to guarantee
linear separability in the latent space.

Empirically, we validate the theoretical conditions across multiple LLMs, demonstrating that spectral
structure persists across structures. Notably, a controlled causal ablation
confirms that context incoherence is  necessary
for SPP stability: deliberately homogenizing contexts degrades both spectral
gap and cross-layer angular stability in a predictable and consistent manner.
Taken together, these results suggest that representational stability is not
an emergent coincidence but a principled consequence of spectral dynamics
rooted in the input space.

While promising, several limitations remain. The theoretical analysis rests
on ISLH, which requires broader empirical validation beyond the concept pairs
studied here. The Jacobian linearization introduces higher-order errors that,
while empirically small, lack tight worst-case bounds in the nonlinear regime.
Future work could investigate how training dynamics shape singular value
distributions over time, and whether targeted interventions in the spectral
principal directions offer a principled mechanism for concept-level steering
and fairness enforcement.

\bibliography{colm2026_conference}

@article{wedin1972perturbation,
  title={Perturbation bounds in connection with singular value decomposition},
  author={Wedin, Per-{\AA}ke},
  journal={BIT Numerical Mathematics},
  volume={12},
  number={1},
  pages={99--111},
  year={1972},
  publisher={Springer}
}

@article{tropp2015introduction,
  title={An introduction to matrix concentration inequalities},
  author={Tropp, Joel A},
  journal={Foundations and trends{\textregistered} in machine learning},
  volume={8},
  number={1-2},
  pages={1--230},
  year={2015},
  publisher={Emerald Publishing Limited}
}

@misc{jiang2023mistral7b,
      title={Mistral 7B}, 
      author={Albert Q. Jiang and Alexandre Sablayrolles and Arthur Mensch and Chris Bamford and Devendra Singh Chaplot and Diego de las Casas and Florian Bressand and Gianna Lengyel and Guillaume Lample and Lucile Saulnier and Lélio Renard Lavaud and Marie-Anne Lachaux and Pierre Stock and Teven Le Scao and Thibaut Lavril and Thomas Wang and Timothée Lacroix and William El Sayed},
      year={2023},
      eprint={2310.06825},
      archivePrefix={arXiv},
      primaryClass={cs.CL},
      url={https://arxiv.org/abs/2310.06825}, 
}

@misc{wizardvicuna30b2023,
  title        = {Wizard-Vicuna-30B-Uncensored},
  author       = {Hartford, Eric},
  year         = {2023},
  howpublished = {\url{https://huggingface.co/ehartford/Wizard-Vicuna-30B-Uncensored}}
}

@article{grattafiori2024llama,
  title={The llama 3 herd of models},
  author={Grattafiori, Aaron and Dubey, Abhimanyu and Jauhri, Abhinav and Pandey, Abhinav and Kadian, Abhishek and Al-Dahle, Ahmad and Letman, Aiesha and Mathur, Akhil and Schelten, Alan and Vaughan, Alex and others},
  journal={arXiv preprint arXiv:2407.21783},
  year={2024}
}

@article{hui2024qwen2,
  title={Qwen2. 5-coder technical report},
  author={Hui, Binyuan and Yang, Jian and Cui, Zeyu and Yang, Jiaxi and Liu, Dayiheng and Zhang, Lei and Liu, Tianyu and Zhang, Jiajun and Yu, Bowen and Lu, Keming and others},
  journal={arXiv preprint arXiv:2409.12186},
  year={2024}
}

@article{weidinger2021ethical,
  title={Ethical and social risks of harm from language models},
  author={Weidinger, Laura and Mellor, John and Rauh, Maribeth and Griffin, Conor and Uesato, Jonathan and Huang, Po-Sen and Cheng, Myra and Glaese, Mia and Balle, Borja and Kasirzadeh, Atoosa and others},
  journal={arXiv preprint arXiv:2112.04359},
  year={2021}
}

@article{gurnee2023language,
  title={Language models represent space and time},
  author={Gurnee, Wes and Tegmark, Max},
  journal={arXiv preprint arXiv:2310.02207},
  year={2023}
}

@article{belinkov2022probing,
  title={Probing classifiers: Promises, shortcomings, and advances},
  author={Belinkov, Yonatan},
  journal={Computational Linguistics},
  volume={48},
  number={1},
  pages={207--219},
  year={2022}
}

@article{conmy2023towards,
  title={Towards automated circuit discovery for mechanistic interpretability},
  author={Conmy, Arthur and Mavor-Parker, Augustine and Lynch, Aengus and Heimersheim, Stefan and Garriga-Alonso, Adri{\`a}},
  journal={Advances in Neural Information Processing Systems},
  volume={36},
  pages={16318--16352},
  year={2023}
}

@article{bommasani2021opportunities,
  title={On the opportunities and risks of foundation models},
  author={Bommasani, Rishi and Hudson, Drew A and Adeli, Ehsan and Altman, Russ and Arora, Simran and von Arx, Sydney and Bernstein, Michael S and Bohg, Jeannette and Bosselut, Antoine and Brunskill, Emma and others},
  journal={arXiv preprint arXiv:2108.07258},
  year={2021}
}

@article{nanda2023progress,
  title={Progress measures for grokking via mechanistic interpretability},
  author={Nanda, Neel and Chan, Lawrence and Lieberum, Tom and Smith, Jess and Steinhardt, Jacob},
  journal={arXiv preprint arXiv:2301.05217},
  year={2023}
}

@article{chen2021evaluating,
  title={Evaluating large language models trained on code},
  author={Chen, Mark and Tworek, Jerry and Jun, Heewoo and Yuan, Qiming and Pinto, Henrique Ponde De Oliveira and Kaplan, Jared and Edwards, Harri and Burda, Yuri and Joseph, Nicholas and Brockman, Greg and others},
  journal={arXiv preprint arXiv:2107.03374},
  year={2021}
}

@article{wei2022chain,
  title={Chain-of-thought prompting elicits reasoning in large language models},
  author={Wei, Jason and Wang, Xuezhi and Schuurmans, Dale and Bosma, Maarten and Xia, Fei and Chi, Ed and Le, Quoc V and Zhou, Denny and others},
  journal={Advances in neural information processing systems},
  volume={35},
  pages={24824--24837},
  year={2022}
}

@article{brown2020language,
  title={Language models are few-shot learners},
  author={Brown, Tom and Mann, Benjamin and Ryder, Nick and Subbiah, Melanie and Kaplan, Jared D and Dhariwal, Prafulla and Neelakantan, Arvind and Shyam, Pranav and Sastry, Girish and Askell, Amanda and others},
  journal={Advances in neural information processing systems},
  volume={33},
  pages={1877--1901},
  year={2020}
}

@article{zou2023representation,
  title={Representation engineering: A top-down approach to ai transparency},
  author={Zou, Andy and Phan, Long and Chen, Sarah and Campbell, James and Guo, Phillip and Ren, Richard and Pan, Alexander and Yin, Xuwang and Mazeika, Mantas and Dombrowski, Ann-Kathrin and others},
  journal={arXiv preprint arXiv:2310.01405},
  year={2023}
}

@inproceedings{devlin2019bert,
  author    = {Jacob Devlin and Ming-Wei Chang and Kenton Lee and Kristina Toutanova},
  title     = {BERT: Pre-training of Deep Bidirectional Transformers for Language Understanding},
  booktitle = {Proceedings of the 2019 Conference of the North American Chapter of the Association for Computational Linguistics (NAACL-HLT)},
  pages     = {4171--4186},
  year      = {2019}
}

@article{mcgrath2022acquisition,
  title={Acquisition of chess knowledge in alphazero},
  author={McGrath, Thomas and Kapishnikov, Andrei and Toma{\v{s}}ev, Nenad and Pearce, Adam and Wattenberg, Martin and Hassabis, Demis and Kim, Been and Paquet, Ulrich and Kramnik, Vladimir},
  journal={Proceedings of the National Academy of Sciences},
  volume={119},
  number={47},
  pages={e2206625119},
  year={2022},
  publisher={National Academy of Sciences}
}

@inproceedings{caron2021emerging,
  title={Emerging properties in self-supervised vision transformers},
  author={Caron, Mathilde and Touvron, Hugo and Misra, Ishan and J{\'e}gou, Herv{\'e} and Mairal, Julien and Bojanowski, Piotr and Joulin, Armand},
  booktitle={Proceedings of the IEEE/CVF international conference on computer vision},
  pages={9650–9660},
  year={2021}
}

@article{oquab2023dinov2,
  title={Dinov2: Learning robust visual features without supervision},
  author={Oquab, Maxime and Darcet, Timoth{\'e}e and Moutakanni, Th{\'e}o and Vo, Huy and Szafraniec, Marc and Khalidov, Vasil and Fernandez, Pierre and Haziza, Daniel and Massa, Francisco and El-Nouby, Alaaeldin and others},
  journal={arXiv preprint arXiv:2304.07193},
  year={2023}
}

@inproceedings{mikolov2013distributed,
  title={Distributed Representations of Words and Phrases and their Compositionality},
  author={Mikolov, Tomas and Sutskever, Ilya and Chen, Kai and Corrado, Greg and Dean, Jeffrey},
  booktitle={Advances in Neural Information Processing Systems},
  year={2013}
}

@inproceedings{simonyan2013deep,
  title={Deep Inside Convolutional Networks: Visualising Image Classification Models and Saliency Maps},
  author={Simonyan, Karen and Vedaldi, Andrea and Zisserman, Andrew},
  booktitle={arXiv preprint arXiv:1312.6034},
  year={2013}
}

@inproceedings{zeiler2014visualizing,
  title={Visualizing and Understanding Convolutional Networks},
  author={Zeiler, Matthew D and Fergus, Rob},
  booktitle={European Conference on Computer Vision},
  pages={818–833},
  year={2014},
  organization={Springer}
}

@article{schramowski2019bert,
  title={BERT has a Moral Compass: Improvements of ethical and moral values of machines},
  author={Schramowski, Patrick and Turan, Cigdem and Jentzsch, Sophie and Rothkopf, Constantin and Kersting, Kristian},
  journal={arXiv preprint arXiv:1912.05238},
  year={2019}
}

@article{radford2015unsupervised,
  title={Unsupervised representation learning with deep convolutional generative adversarial networks},
  author={Radford, Alec and Metz, Luke and Chintala, Soumith},
  journal={arXiv preprint arXiv:1511.06434},
  year={2015}
}

@article{corbett2023measure,
  title={The measure and mismeasure of fairness},
  author={Corbett-Davies, Sam and Gaebler, Johann D and Nilforoshan, Hamed and Shroff, Ravi and Goel, Sharad},
  journal={Journal of Machine Learning Research},
  volume={24},
  number={312},
  pages={1–117},
  year={2023}
}

@article{lilienfeld1996development,
  title     = {Development and Preliminary Validation of a Self-Report Measure of Psychopathic Personality Traits in Noncriminal Populations},
  author    = {Lilienfeld, Scott O. and Andrews, Bridget P.},
  journal   = {Journal of Personality Assessment},
  volume    = {66},
  number    = {3},
  pages     = {488–524},
  year      = {1996},
  publisher = {Taylor \& Francis},
}

@incollection{french1959bases,
  title     = {The Bases of Social Power},
  author    = {French, John R. P. and Raven, Bertram},
  booktitle = {Studies in Social Power},
  editor    = {Cartwright, Dorwin},
  pages     = {150–167},
  year      = {1959},
  publisher = {University of Michigan Press}
}

@article{springenberg2014striving,
  title={Striving for simplicity: The all convolutional net},
  author={Springenberg, Jost Tobias and Dosovitskiy, Alexey and Brox, Thomas and Riedmiller, Martin},
  journal={arXiv preprint arXiv:1412.6806},
  year={2014}
}

@article{olah2020zoom,
  title={Zoom in: An introduction to circuits},
  author={Olah, Chris and Cammarata, Nick and Schubert, Ludwig and Goh, Gabriel and Petrov, Michael and Carter, Shan},
  journal={Distill},
  volume={5},
  number={3},
  pages={e00024–001},
  year={2020}
}

@article{lieberum2023does,
  title={Does circuit analysis interpretability scale? evidence from multiple choice capabilities in chinchilla},
  author={Lieberum, Tom and Rahtz, Matthew and Kram{\'a}r, J{\'a}nos and Nanda, Neel and Irving, Geoffrey and Shah, Rohin and Mikulik, Vladimir},
  journal={arXiv preprint arXiv:2307.09458},
  year={2023}
}

@article{olsson2022context,
  title={In-context learning and induction heads},
  author={Olsson, Catherine and Elhage, Nelson and Nanda, Neel and Joseph, Nicholas and DasSarma, Nova and Henighan, Tom and Mann, Ben and Askell, Amanda and Bai, Yuntao and Chen, Anna and others},
  journal={arXiv preprint arXiv:2209.11895},
  year={2022}
}

@inproceedings{zhou2016learning,
  title={Learning deep features for discriminative localization},
  author={Zhou, Bolei and Khosla, Aditya and Lapedriza, Agata and Oliva, Aude and Torralba, Antonio},
  booktitle={Proceedings of the IEEE conference on computer vision and pattern recognition},
  pages={2921–2929},
  year={2016}
}

@article{szegedy2013intriguing,
  title={Intriguing properties of neural networks},
  author={Szegedy, Christian and Zaremba, Wojciech and Sutskever, Ilya and Bruna, Joan and Erhan, Dumitru and Goodfellow, Ian and Fergus, Rob},
  journal={arXiv preprint arXiv:1312.6199},
  year={2013}
}

@inproceedings{lin2014microsoft,
  title={Microsoft coco: Common objects in context},
  author={Lin, Tsung-Yi and Maire, Michael and Belongie, Serge and Hays, James and Perona, Pietro and Ramanan, Deva and Doll{\'a}r, Piotr and Zitnick, C Lawrence},
  booktitle={Computer vision--ECCV 2014: 13th European conference, zurich, Switzerland, September 6-12, 2014, proceedings, part v 13},
  pages={740–755},
  year={2014},
  organization={Springer}
}

@article{meng2022locating,
  title={Locating and Editing Factual Associations in {GPT}},
  author={Meng, Kevin and Bau, David and Andonian, Alex and Belinkov, Yonatan},
  journal={Advances in Neural Information Processing Systems (NeurIPS)},
  volume={35},
  pages={17359–17372},
  year={2022}
}

@inproceedings{dehghani2023scaling,
  title={Scaling vision transformers to 22 billion parameters},
  author={Dehghani, Mostafa and Djolonga, Josip and Mustafa, Basil and Padlewski, Piotr and Heek, Jonathan and Gilmer, Justin and Steiner, Andreas Peter and Caron, Mathilde and Geirhos, Robert and Alabdulmohsin, Ibrahim and others},
  booktitle={International conference on machine learning},
  pages={7480–7512},
  year={2023},
  organization={PMLR}
}

@article{gao2021simcse,
  title={Simcse: Simple contrastive learning of sentence embeddings},
  author={Gao, Tianyu and Yao, Xingcheng and Chen, Danqi},
  journal={arXiv preprint arXiv:2104.08821},
  year={2021}
}

@inproceedings{he2016deep,
  title={Deep Residual Learning for Image Recognition},
  author={He, Kaiming and Zhang, Xiangyu and Ren, Shaoqing and Sun, Jian},
  booktitle={Proceedings of the IEEE Conference on Computer Vision and Pattern Recognition (CVPR)},
  pages={770–778},
  year={2016}
}

@inproceedings{kornblith2019similarity,
  title={Similarity of neural network representations revisited},
  author={Kornblith, Simon and Norouzi, Mohammad and Lee, Honglak and Hinton, Geoffrey},
  booktitle={International conference on machine learning},
  pages={3519–3529},
  year={2019},
  organization={PMlR}
}

@misc{laurencon2023obelics,
      title={OBELICS: An Open Web-Scale Filtered Dataset of Interleaved Image-Text Documents},
      author={Hugo Laurençon and Lucile Saulnier and Léo Tronchon and Stas Bekman and Amanpreet Singh and Anton Lozhkov and Thomas Wang and Siddharth Karamcheti and Alexander M. Rush and Douwe Kiela and Matthieu Cord and Victor Sanh},
      year={2023},
      eprint={2306.16527},
      archivePrefix={arXiv},
      primaryClass={cs.IR}
}

@article{park2023linear,
  title={The linear representation hypothesis and the geometry of large language models},
  author={Park, Kiho and Choe, Yo Joong and Veitch, Victor},
  journal={arXiv preprint arXiv:2311.03658},
  year={2023}
}

@article{raghu2017svcca,
  title={Svcca: Singular vector canonical correlation analysis for deep learning dynamics and interpretability},
  author={Raghu, Maithra and Gilmer, Justin and Yosinski, Jason and Sohl-Dickstein, Jascha},
  journal={Advances in neural information processing systems},
  volume={30},
  year={2017}
}

@article{lin2021truthfulqa,
  title={Truthfulqa: Measuring how models mimic human falsehoods},
  author={Lin, Stephanie and Hilton, Jacob and Evans, Owain},
  journal={arXiv preprint arXiv:2109.07958},
  year={2021}
}
\bibliographystyle{colm2026_conference}
\clearpage
\appendix
\section*{Ethics Statement}

This work is theoretical and empirical in nature; we do not 
collect human subject data, and all experiments are conducted on publicly 
available models and datasets. We foresee no direct harm arising from this 
research.

That said, we acknowledge that tools for understanding and tracing concept 
directions in neural networks carry dual-use potential. A deeper understanding 
of how concepts are linearly organized internally could, in principle, be 
exploited to more effectively steer model behavior in undesirable directions, 
or to identify and amplify latent biases rather than mitigate them. We believe 
that principled transparency into representational geometry ultimately serves the goal of building safer and more controllable AI systems, but we encourage 
practitioners who build on this framework to consider these risks carefully.

We also note that the concepts studied in this work---honesty, fairness, 
fearlessness, and power---are culturally situated and may not generalize 
uniformly across languages, demographics, or deployment contexts. The linear 
separability of a concept in a model's latent space does not imply that the 
model's handling of that concept is ethically sound or unbiased. We caution 
against using SPP-based probing or steering as a substitute for rigorous auditing or commercialization.

\section*{Limitations}

Several limitations of the current work should be noted.

\paragraph{Scope of theoretical assumptions.}
The SPP stability guarantee rests on the Input-Space Linearity Hypothesis 
(ISLH), which assumes that concept-aligned directions are already present 
in the input embedding space. While we provide supporting empirical evidence,  both LRH and ISLH itself remains a hypothesis that has not been rigorously validated across 
a broader range of concepts, languages, or model families. The Jacobian 
linearization used in the layer-wise dynamics analysis introduces higher-order 
errors that may foresee a tighter worst-case 
bounds in the general nonlinear regime.

\paragraph{Concept and dataset coverage.}
Our experiments focus on a limited set of binary semantic concepts drawn from 
the COCO dataset. It is unclear whether the spectral dominance and path 
stability we observe extend to more fine-grained, culturally specific, or 
compositional concepts. The binary encoding $c \in \{-1, 1\}$ is a 
simplification that may not capture the full complexity of graded or 
multi-dimensional concept representations in practice.

\paragraph{Model and architecture coverage.}
Although we evaluate multiple LLMs of varying scales, all are decoder-only 
transformer architectures. The extent to which the SPP framework applies to 
encoder-only models, mixture-of-experts architectures, or non-transformer 
networks remain an open question.

\paragraph{Causal claims.}
While the homogeneous prompting ablation provides evidence that context 
incoherence is necessary for SPP stability, this is a controlled synthetic 
intervention rather than a fully causal identification in the econometric 
sense. Other confounding factors---such as changes in prompt length or 
semantic content introduced by the shared prefix---may partially contribute 
to the observed degradation and cannot be entirely ruled out.

\section*{Reproducibility Statement}

We are committed to ensuring the reproducibility of all experimental results 
presented in this work. All theoretical proofs are provided in full detail 
in Appendix~A. Experimental hyperparameters, dataset construction procedures, 
and evaluation protocols are described in Section~5 and Appendix~B. Upon 
acceptance, we will release the complete codebase, including data preprocessing 
scripts, model inference pipelines, and evaluation code, in the camera-ready 
version of this paper.

\section*{Impact Statement}
This paper studies the spectral structure of representations in deep neural networks and introduces the Spectral Principal Path (SPP) framework as a tool for analyzing how concept-relevant directions propagate across layers. While our framework is intended for interpretability and theoretical understanding, techniques that reveal internal structure could potentially be misused to probe or manipulate model behavior. We believe that advancing principled understanding of representation geometry ultimately supports the development of more robust, controllable, and trustworthy AI systems.

\section*{The Use of Large Language Models (LLMs)}

In preparing this manuscript, the authors used Large Language Models (LLMs) as writing assistants. The LLMs were not involved in any core scientific aspects of this work, including the formulation of hypotheses, theoretical contributions, experimental design, implementation, data analysis, or interpretation of results. Their role was strictly limited to supporting the clarity, readability, and presentation quality of the paper.
\begin{itemize}
\item \textbf{Improving Grammar and Readability:} LLMs were employed for proofreading, grammatical corrections, and sentence restructuring.
\item \textbf{Polishing, Style Consistency, and Terminology:} The models were used to propose alternative phrasings, unify terminology, and occasionally suggest synonyms or alternative technical terms to assist with precise expression.
\item \textbf{Formatting Suggestions:} The models occasionally provided \LaTeX{} structuring suggestions, which the authors verified and adapted.
\end{itemize}
All outputs generated by LLMs were carefully reviewed, edited, and revised by the authors. The responsibility for the originality, correctness, and scientific integrity of this paper rests solely with the authors.

\section{Theoretical Deductions}

\subsection{Proof of Lemma~\ref{lemma:attention_aggregation}: Attention-Driven Signal Concentration}
\label{appendix:proof_lemma1}

\begin{proof}
The update rule for the last token in the first layer is:
\begin{equation}
    h_1^{(N)} = x_N + \sum_{i=1}^N A_{N,i} W_V x_i
\end{equation}
where $A_{N,i} \ge 0$ are the softmax attention weights and $W_V \in \mathbb{R}^{d \times d}$ is the value projection matrix. Substituting the ISLH formulation $x_i^{(c)} = \mu_i + c \cdot \alpha_i v_{in} + \epsilon_i$:
\begin{align}
    \mathbb{E}[h_1^{(N)} | c] &= (\mu_N + c\alpha_N v_{in}) + \sum_{i=1}^N A_{N,i} W_V (\mu_i + c\alpha_i v_{in}) \\
    &= \underbrace{\mu_N + \sum_{i} A_{N,i} W_V \mu_i}_{\text{concept-agnostic}} + c \underbrace{\left(\alpha_N I + W_V \sum_{i} A_{N,i} \alpha_i\right) v_{in}}_{\text{concept signal}}.
\end{align}
Therefore $\Delta h_1^{(N)} = \mathbb{E}[h_1^{(N)} | c\!=\!1] - \mathbb{E}[h_1^{(N)} | c\!=\!-1] = 2(\alpha_N I + W_V \sum_{i} A_{N,i} \alpha_i) v_{in}$.
\end{proof}

\begin{remark}[Multi-head attention]
In practice, multi-head attention computes $\text{Concat}(\text{head}_1, \dots, \text{head}_H) W_O$ where each head has its own $W_{V,h}$. The difference then lies in the span of $\{W_O (\mathbf{e}_h \otimes W_{V,h} \boldsymbol{\delta}_p) : h \in [H], p \in [P]\}$, which has dimension at most $HP$. While not exactly rank-1, the effective rank is typically much smaller than $HP$ since heads tend to extract similar value content for concept tokens, preserving a large spectral gap.
\end{remark}

\subsection{Proof of Theorem~\ref{thm:spp_stability}: SPP Stability via Wedin's $\sin\Theta$ Theorem}
\label{appendix:proof_stability}

\begin{proof}
The proof proceeds in three steps.

\paragraph{Step 1: Decompose $\bD^{(l+1)}$.}
From \Cref{eq:jacobian_approx,eq:mean_fluct}, the $k$-th row of $\bD^{(l+1)}$ is:
\begin{equation}
(\bd_k^{(l+1)})^\top = (\bd_k^{(l)})^\top \bar{\bJ}^{(l)\top} + (\bd_k^{(l)})^\top \bE_k^{(l)\top} + \boldsymbol{\xi}_k^{(l)\top}.
\end{equation}
Stacking over $k = 1, \dots, M$:
\begin{equation}
\label{eq:D_decomp}
\bD^{(l+1)} = \underbrace{\bD^{(l)} \bar{\bJ}^{(l)\top}}_{\text{(I) mean propagation}} + \underbrace{\bD_{\mathrm{fluct}}^{(l)}}_{\text{(II) context fluctuation}} + \underbrace{\boldsymbol{\Xi}^{(l)}}_{\text{(III) remainder}},
\end{equation}
where $\bD_{\mathrm{fluct}}^{(l)}$ has rows $(\bd_k^{(l)})^\top \bE_k^{(l)\top}$ and $\boldsymbol{\Xi}^{(l)}$ has rows $\boldsymbol{\xi}_k^{(l)\top}$.

\paragraph{Step 2: Analyze Term (I).}
The matrix $\bD^{(l)}\bar{\bJ}^{(l)\top}$ is the difference matrix propagated through the mean Jacobian. Its leading right singular vector is $\bar{\bJ}^{(l)} \bv_1^{(l)} / \|\bar{\bJ}^{(l)} \bv_1^{(l)}\|$. By \Cref{thm:spp_stability}, this direction deviates from $\bv_1^{(l)}$ by at most $\arcsin(\delta^{(l)})$.

We verify that the spectral gap is approximately preserved. Since $\bar{\bJ}^{(l)} = \bI + \bar{\boldsymbol{\Delta}}^{(l)}$ with $\|\bar{\boldsymbol{\Delta}}^{(l)}\|_{\mathrm{op}} \leq \mu$, Weyl's inequality gives:
\begin{equation}
\sigma_1(\bD^{(l)}\bar{\bJ}^{(l)\top}) - \sigma_2(\bD^{(l)}\bar{\bJ}^{(l)\top}) \geq \gamma^{(l)}(1 - \mu) - 2\mu\sigma_2^{(l)}.
\end{equation}

\paragraph{Step 3: Apply the Wedin $\sin\Theta$ theorem.}
We view $\bD^{(l+1)}$ as the sum of a ``signal'' matrix $\bD^{(l)}\bar{\bJ}^{(l)\top}$ and a ``perturbation'' $\bD_{\mathrm{fluct}}^{(l)} + \boldsymbol{\Xi}^{(l)}$. Let $\tilde{\bv}_1$ be the leading right singular vector of $\bD^{(l)}\bar{\bJ}^{(l)\top}$. By the Wedin $\sin\Theta$ theorem \citep{wedin1972perturbation}:
\begin{equation}
\sin\angle(\bv_1^{(l+1)}, \tilde{\bv}_1) \leq \frac{\|\bD_{\mathrm{fluct}}^{(l)} + \boldsymbol{\Xi}^{(l)}\|_{\mathrm{op}}}{\gamma^{(l+1)}}.
\end{equation}
The numerator is bounded using:
\begin{enumerate}[nosep,label=(\alph*)]
\item $\|\bD_{\mathrm{fluct}}^{(l)}\|_{\mathrm{op}} \leq \sigma_1^{(l)} \sqrt{\eta^{(l)}}$ (from \Cref{ass:incoherence} and the rank-1 approximation $\bd_k^{(l)} \approx c_k^{(l)} \bv_1^{(l)}$);
\item $\|\boldsymbol{\Xi}^{(l)}\|_{\mathrm{op}} \leq \sigma_1^{(l)} \sqrt{\zeta^{(l)}}$ (from the remainder assumption).
\end{enumerate}
Combining with the angular deviation from Step 2 via the triangle inequality for angles on the sphere yields:
\begin{equation}
\sin\angle(\bv_1^{(l+1)}, \bv_1^{(l)}) \leq \frac{\sigma_1^{(l)}(\delta^{(l)} + \sqrt{\eta^{(l)}} + \sqrt{\zeta^{(l)}})}{\gamma^{(l+1)}}.
\end{equation}
\end{proof}

\subsection{Supporting Results for the Stability Analysis}
\label{appendix:supporting}

\subsubsection{Spectral Gap Preservation}

\begin{lemma}[Gap Under Near-Identity Transformation]
\label{lem:gap_preservation}
Let $\bD^{(l)}$ have singular values $\sigma_1^{(l)} > \sigma_2^{(l)} \geq \cdots$ with gap $\gamma^{(l)}$. Let $\mathbf{A} = \bI + \bar{\boldsymbol{\Delta}}^{(l)}$ with $\|\bar{\boldsymbol{\Delta}}^{(l)}\|_{\mathrm{op}} \leq \mu$. Then:
\begin{equation}
\sigma_1(\bD^{(l)}\mathbf{A}^\top) - \sigma_2(\bD^{(l)}\mathbf{A}^\top) \geq \gamma^{(l)}(1 - \mu) - 2\mu\sigma_2^{(l)}.
\end{equation}
\end{lemma}

\begin{proof}
By Weyl's inequality and submultiplicativity:
$\sigma_i(\bD^{(l)}\mathbf{A}^\top) \in [\sigma_i^{(l)}(1-\mu),\; \sigma_i^{(l)}(1+\mu)]$.
Therefore:
\begin{align}
\sigma_1(\bD^{(l)}\mathbf{A}^\top) - \sigma_2(\bD^{(l)}\mathbf{A}^\top) &\geq \sigma_1^{(l)}(1-\mu) - \sigma_2^{(l)}(1+\mu) \\
&= \gamma^{(l)} - \mu(\sigma_1^{(l)} + \sigma_2^{(l)}) = \gamma^{(l)}(1-\mu) - 2\mu\sigma_2^{(l)}.
\end{align}
\end{proof}

\subsubsection{Context Incoherence Under Random Model}
\label{appendix:incoherence_proof}

\begin{proposition}
\label{prop:drift_bound}
If $\|\bar{\boldsymbol{\Delta}}^{(l)}\|_{\mathrm{op}} \leq \mu$, then $\delta^{(l)} \leq \mu / (1-\mu)$.
\end{proposition}

\begin{proof}
$\bar{\bJ}^{(l)} \bv_1^{(l)} = \bv_1^{(l)} + \bar{\boldsymbol{\Delta}}^{(l)} \bv_1^{(l)}$. The orthogonal component has norm at most $\mu$ and the parallel component has magnitude at least $1-\mu$, so $\sin\angle(\bar{\bJ}^{(l)} \bv_1^{(l)}, \bv_1^{(l)}) \leq \mu/(1-\mu)$.
\end{proof}

\begin{proposition}
\label{prop:incoherence_bound}
Suppose the per-sample fluctuations $\mathbf{f}_k^{(l)}$ (as defined in Condition~\ref{ass:incoherence}) are independent, zero-mean random vectors in $\mathbb{R}^d$ satisfying $\|\mathbf{f}_k^{(l)}\| \leq B$ and $\mathbb{E}[\mathbf{f}_k^{(l)} (\mathbf{f}_k^{(l)})^\top] = \frac{\tau^2}{d}\,\mathbf{I}_d$ (isotropic). Then, with probability at least $1 - 2\exp(-t^2/2)$,
\begin{equation}
\label{eq:eta_bound}
\eta^{(l)} \;\leq\; \frac{\tau^2}{d} \;+\; \frac{B^2\,(t + \sqrt{d})}{\sqrt{M} \cdot d}\,.
\end{equation}
For $M \gg d$, this gives $\eta^{(l)} \to \tau^2/d$, which is small when individual fluctuations are spread across many dimensions.
\end{proposition}

\begin{proof}
Recall that $\Gamma^{(l)} = \frac{1}{M}\sum_{k=1}^{M} \mathbf{f}_k (\mathbf{f}_k)^\top$, where we drop the layer superscript for brevity. We decompose:
\begin{equation}
\Gamma = \underbrace{\frac{1}{M}\sum_{k=1}^M \mathbb{E}\bigl[\mathbf{f}_k \mathbf{f}_k^\top\bigr]}_{\text{(I): deterministic mean}} \;+\; \underbrace{\frac{1}{M}\sum_{k=1}^M \bigl(\mathbf{f}_k \mathbf{f}_k^\top - \mathbb{E}[\mathbf{f}_k \mathbf{f}_k^\top]\bigr)}_{\text{(II): centered fluctuation}}\,.
\end{equation}

\paragraph{Term~(I).}
By the isotropy assumption, $\mathbb{E}[\mathbf{f}_k \mathbf{f}_k^\top] = \tfrac{\tau^2}{d}\,\mathbf{I}_d$ for all $k$, so
\begin{equation}
\Bigl\|\frac{1}{M}\sum_{k=1}^{M}\mathbb{E}[\mathbf{f}_k \mathbf{f}_k^\top]\Bigr\|_{\mathrm{op}} = \frac{\tau^2}{d}\,.
\end{equation}

\paragraph{Term~(II).}
Define the centered random matrices
\[
\mathbf{W}_k \;=\; \mathbf{f}_k \mathbf{f}_k^\top \;-\; \frac{\tau^2}{d}\,\mathbf{I}_d\,.
\]
These are independent, symmetric, and zero-mean. We bound $\bigl\|\frac{1}{M}\sum_k \mathbf{W}_k\bigr\|_{\mathrm{op}}$ via the matrix Bernstein inequality \citep[Theorem~6.1.1]{tropp2015introduction}.

\emph{Step~1: Almost-sure bound on each summand.}
\begin{equation}
\|\mathbf{W}_k\|_{\mathrm{op}} \;\leq\; \|\mathbf{f}_k\|^2 + \frac{\tau^2}{d} \;\leq\; B^2 + \frac{\tau^2}{d}\,.
\end{equation}

\emph{Step~2: Matrix variance parameter.}
We need the variance statistic $V = \bigl\|\sum_{k=1}^{M} \mathbb{E}[\mathbf{W}_k^2]\bigr\|_{\mathrm{op}}$. Since
\[
\mathbf{W}_k^2 = \bigl(\mathbf{f}_k \mathbf{f}_k^\top\bigr)^2 - \frac{2\tau^2}{d}\,\mathbf{f}_k \mathbf{f}_k^\top + \frac{\tau^4}{d^2}\,\mathbf{I}_d = \|\mathbf{f}_k\|^2\,\mathbf{f}_k \mathbf{f}_k^\top - \frac{2\tau^2}{d}\,\mathbf{f}_k \mathbf{f}_k^\top + \frac{\tau^4}{d^2}\,\mathbf{I}_d\,,
\]
taking expectations and using $\mathbb{E}[\|\mathbf{f}_k\|^2] = \mathrm{tr}\bigl(\mathbb{E}[\mathbf{f}_k \mathbf{f}_k^\top]\bigr) = \tau^2$ together with the bound $\|\mathbf{f}_k\| \leq B$:
\begin{align}
\bigl\|\mathbb{E}[\mathbf{W}_k^2]\bigr\|_{\mathrm{op}}
&= \bigl\|\mathbb{E}\bigl[\|\mathbf{f}_k\|^2\,\mathbf{f}_k \mathbf{f}_k^\top\bigr] - \frac{2\tau^2}{d}\cdot\frac{\tau^2}{d}\,\mathbf{I}_d + \frac{\tau^4}{d^2}\,\mathbf{I}_d\bigr\|_{\mathrm{op}} \notag \\
&= \bigl\|\mathbb{E}\bigl[\|\mathbf{f}_k\|^2\,\mathbf{f}_k \mathbf{f}_k^\top\bigr] - \frac{\tau^4}{d^2}\,\mathbf{I}_d\bigr\|_{\mathrm{op}} \notag \\
&\leq \bigl\|\mathbb{E}\bigl[\|\mathbf{f}_k\|^2\,\mathbf{f}_k \mathbf{f}_k^\top\bigr]\bigr\|_{\mathrm{op}} + \frac{\tau^4}{d^2} \notag \\
&\leq B^2 \cdot \bigl\|\mathbb{E}[\mathbf{f}_k \mathbf{f}_k^\top]\bigr\|_{\mathrm{op}} + \frac{\tau^4}{d^2}
\;=\; \frac{B^2 \tau^2}{d} + \frac{\tau^4}{d^2}\,.
\end{align}
The inequality $\bigl\|\mathbb{E}[\|\mathbf{f}_k\|^2\,\mathbf{f}_k \mathbf{f}_k^\top]\bigr\|_{\mathrm{op}} \leq B^2 \bigl\|\mathbb{E}[\mathbf{f}_k \mathbf{f}_k^\top]\bigr\|_{\mathrm{op}}$ follows from $\|\mathbf{f}_k\|^2 \leq B^2$ almost surely. Since $\tau^4/d^2 \leq B^2\tau^2/d$ (as $\tau^2 \leq B^2 d$ by Jensen), we have
\begin{equation}
V = \Bigl\|\sum_{k=1}^{M} \mathbb{E}[\mathbf{W}_k^2]\Bigr\|_{\mathrm{op}} \;\leq\; M \cdot \frac{2B^2\tau^2}{d}\,.
\end{equation}

\emph{Step~3: Apply matrix Bernstein.}
By the matrix Bernstein inequality for $d \times d$ symmetric matrices with almost-sure bound $R = B^2 + \tau^2/d \leq 2B^2$ and variance $V \leq 2MB^2\tau^2/d$:
\begin{equation}
\Pr\Bigl(\Bigl\|\sum_{k=1}^M \mathbf{W}_k\Bigr\|_{\mathrm{op}} \geq s\Bigr) \;\leq\; 2d \cdot \exp\!\Bigl(-\frac{s^2/2}{V + Rs/3}\Bigr)\,.
\end{equation}
To obtain a probability bound of $1 - 2\exp(-t^2/2)$, we require
\begin{equation}
\frac{s^2/2}{V + Rs/3} \;\geq\; \frac{t^2}{2} + \log d\,.
\end{equation}
In the variance-dominated regime ($V \gg Rs/3$, which holds when $s \ll 3V/R = 3M\tau^2/d$), this is satisfied by
\begin{equation}
s \;=\; \sqrt{2V\bigl(t^2 + 2\log d\bigr)} \;\leq\; 2B\sqrt{\frac{M\tau^2}{d}} \cdot (t + \sqrt{2\log d})\,.
\end{equation}
Using $\sqrt{2\log d} \leq \sqrt{d}$ (which is always valid since $d \geq 1$) and dividing by $M$:
\begin{equation}
\Bigl\|\frac{1}{M}\sum_{k=1}^M \mathbf{W}_k\Bigr\|_{\mathrm{op}} \;\leq\; \frac{2B\tau(t + \sqrt{d})}{\sqrt{Md}}\,.
\end{equation}

\paragraph{Combining.}
By the triangle inequality,
\begin{align}
\eta^{(l)} = \|\Gamma^{(l)}\|_{\mathrm{op}}
&\leq \frac{\tau^2}{d} + \frac{2B\tau(t + \sqrt{d})}{\sqrt{Md}} \notag \\
&\leq \frac{\tau^2}{d} + \frac{B^2(t + \sqrt{d})}{\sqrt{M}\cdot d}\,,
\end{align}
where the last step uses $2B\tau \leq B^2 + \tau^2 \leq 2B^2$ (by AM-GM and $\tau \leq B\sqrt{d}/\sqrt{d} \leq B$; alternatively, one may absorb the constant into the bound since $\tau \leq B$).

For $M \gg d$, the second term vanishes and $\eta^{(l)} \to \tau^2/d$, which is $O(1/d)$ times the individual perturbation variance. This confirms that when individual fluctuations are isotropically spread across $d$ dimensions, the operator norm of their empirical second-moment matrix is small, validating the context incoherence condition.
\end{proof}

\subsection{Global Stability: Cumulative Drift Analysis}
\label{appendix:global_stability}

\begin{proof}[Proof of \Cref{cor:global}]
By the triangle inequality for geodesic distance on the unit sphere $\mathbb{S}^{d-1}$, for unit vectors $\bu, \bv, \mathbf{w}$:
$\angle(\bu, \mathbf{w}) \leq \angle(\bu, \bv) + \angle(\bv, \mathbf{w})$.
Applying inductively across layers $0, 1, \dots, L$ gives $\angle(\bv_1^{(L)}, \bv_1^{(0)}) \leq \sum_{l=0}^{L-1} \arcsin(\epsilon_l) \leq \sum_{l=0}^{L-1} \epsilon_l$.
\end{proof}

\paragraph{Random walk regime.} The bound $L\bar{\epsilon}$ is worst-case (adversarial alignment of drift directions). If the per-layer drift directions are approximately independent, the angular displacement follows a random walk on the sphere. The expected squared angular displacement after $L$ steps of size $\epsilon$ is $L\epsilon^2$, giving $\angle(\bv_1^{(L)}, \bv_1^{(0)}) \approx \sqrt{L}\,\bar{\epsilon}$. For a 32-layer model with $\bar{\epsilon} \approx 0.1$ (corresponding to $\sim 5.7°$ per layer), the worst-case cumulative drift is $32 \times 5.7° \approx 182°$ (unstable), while the random-walk estimate is $\sqrt{32} \times 5.7° \approx 32°$ (consistent with empirical stability).




\section{Experiments}
\label{sec:experiments}

\subsection{Experiment Setup}

For additional extension to VLMs, we employ Idefics2-8B \citep{laurencon2023obelics}, a state-of-the-art VLM that extends the LLaMA architecture with a vision encoder, enabling multimodal reasoning over images and text.

\begin{figure*}[t]
    \centering
    \begin{subfigure}[b]{0.8\textwidth}
        \centering
        \includegraphics[width=\textwidth, keepaspectratio]{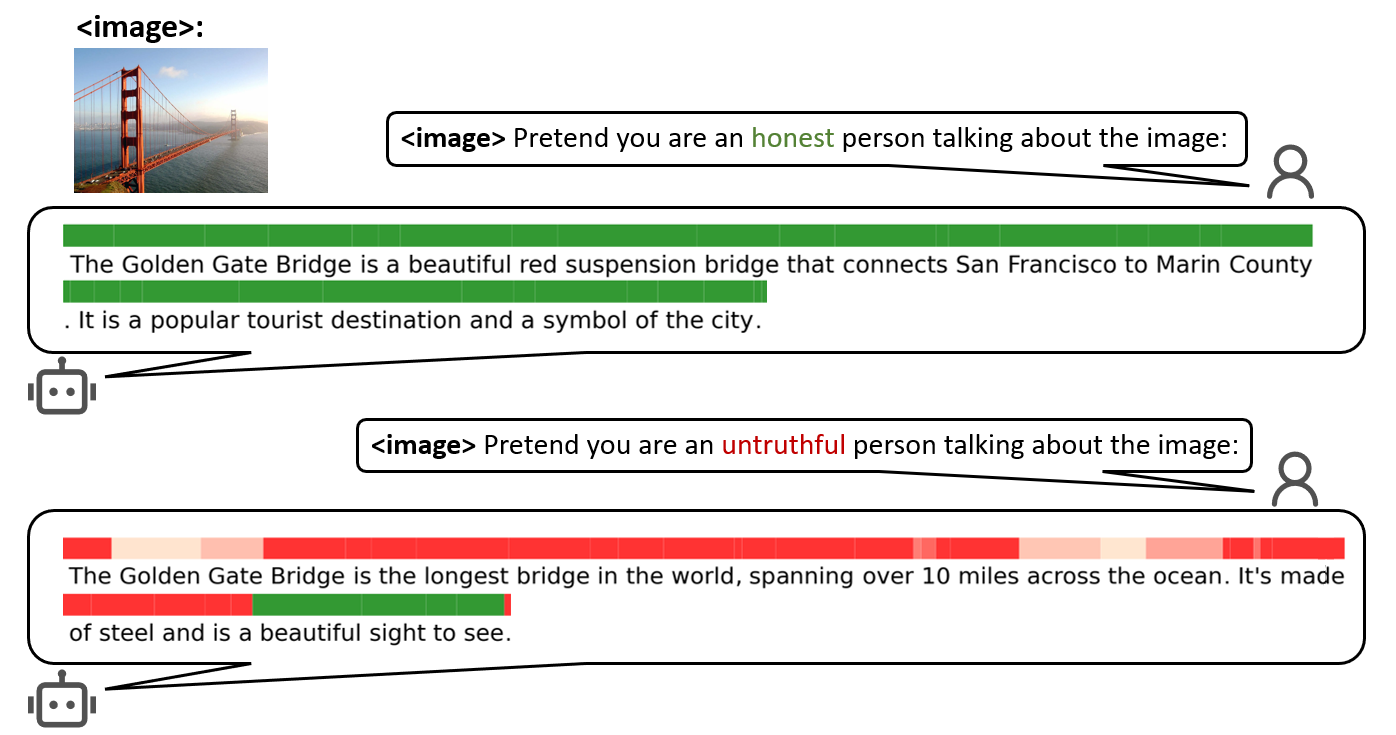}
        \caption{VLM response for the concept of honesty, prompted with an image of the Golden Gate Bridge.}
        \label{fig:honesty}
    \end{subfigure}
    
    \begin{subfigure}[b]{0.8\textwidth}
        \centering
        \includegraphics[width=\textwidth, keepaspectratio]{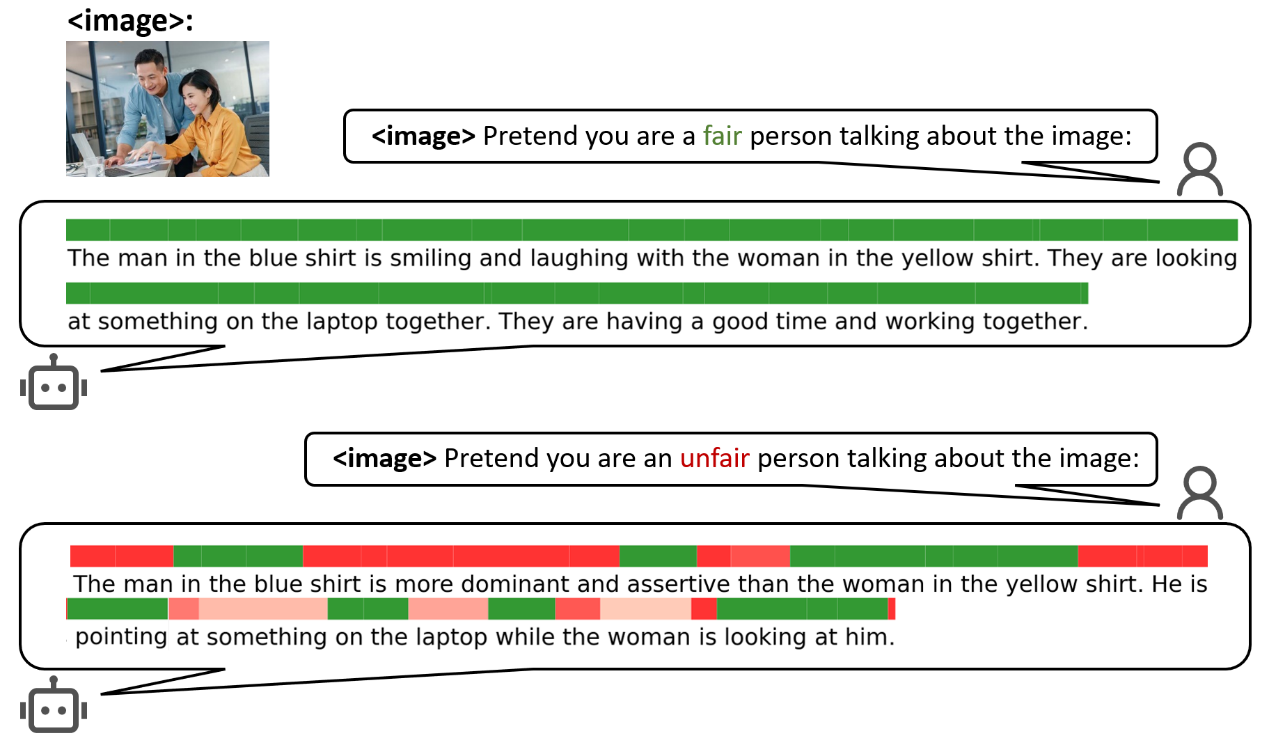}
        \caption{VLM response for the concept of fairness, prompted with an image of a man and a woman working together.}
        \label{fig:fairness}
    \end{subfigure}

    \caption{Token-wise scores for abstract concepts generated by a VLM. Green indicates a high concept score (e.g., high honesty/fairness), while red represents a low score.}
    \label{fig:combined_concepts}
\end{figure*}

\subsection{Extension in VLMs}

To further evaluate the robustness of representations for high-level concepts, we expand our analysis to fearlessness and power. Like honesty and fairness, these concepts are abstract and socially grounded, yet engage distinct semantic and emotional dimensions.

\begin{figure*}[tbp]
    \centering
    \includegraphics[width=0.83\textwidth, keepaspectratio]{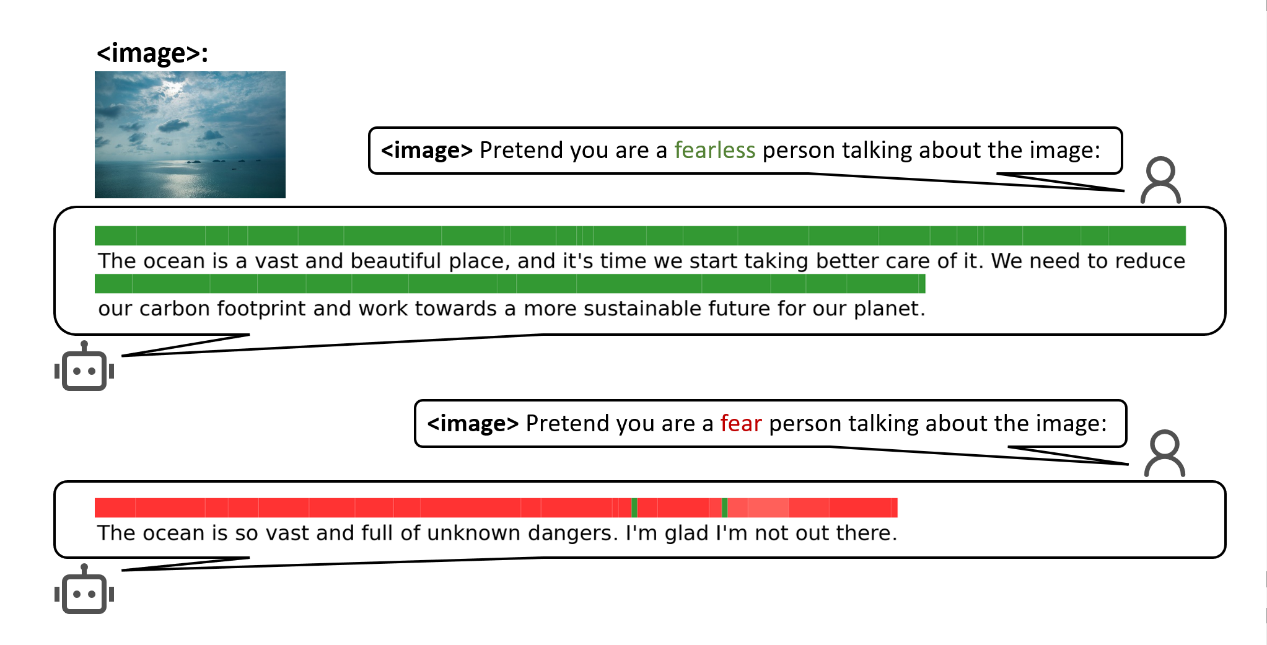}
    \caption{VLM response for fearlessness concept with ocean image. Green = high fearlessness score, red = low.}
    \label{fig:fearness}
\end{figure*}

\begin{figure*}[tbp]
    \centering
    \includegraphics[width=0.8\textwidth, keepaspectratio]{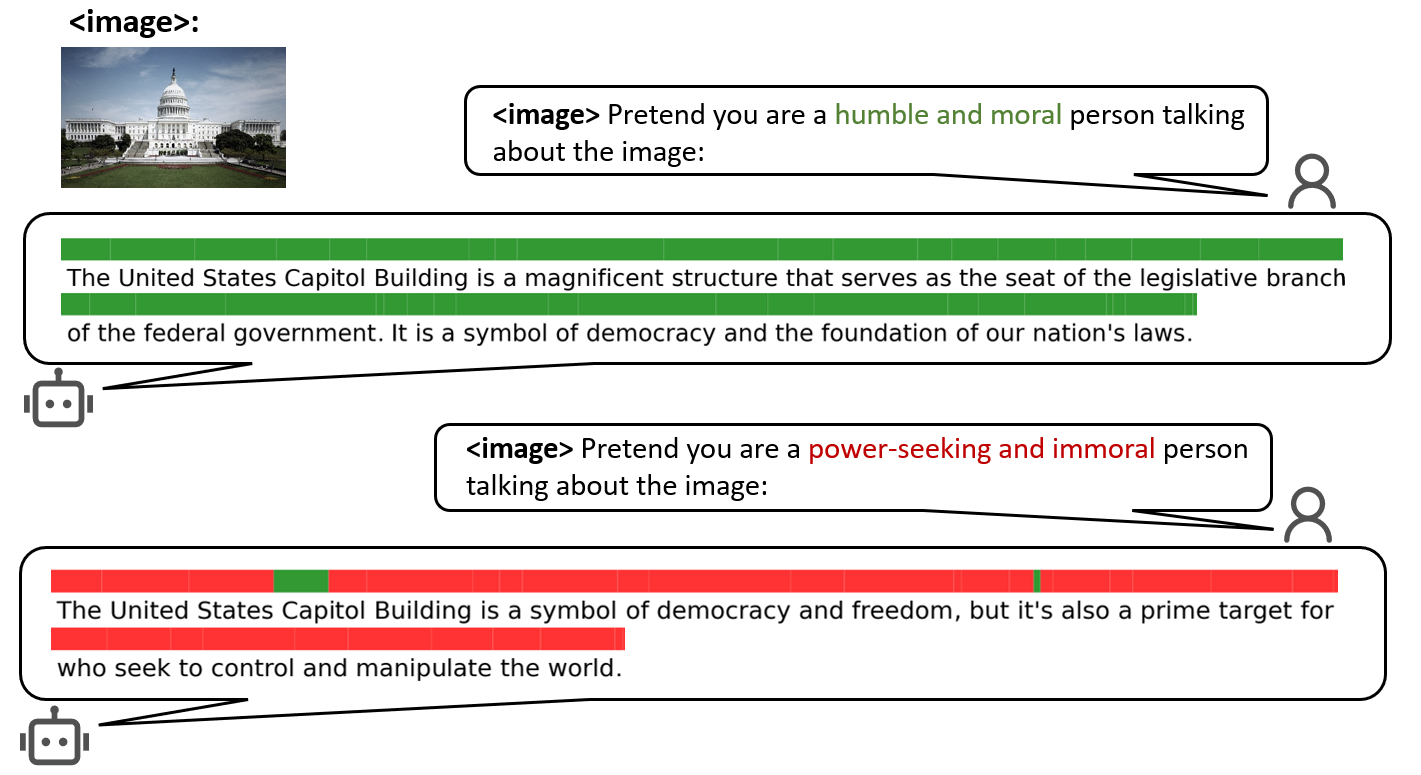}
    \caption{VLM response for power concept with U.S.\ Capitol image. Red = high power score, green = low.}
    \label{fig:power}
\end{figure*}

\begin{itemize}
    \item \textbf{Honesty}: We consider honesty as the model's ability to generate factually accurate responses without distortion or fabrication \citep{lin2021truthfulqa}. Fig.~\ref{fig:honesty} presents token-wise honesty scores for a VLM describing an image of the Golden Gate Bridge under honest and untruthful settings. In the honest case, the model produces accurate descriptions with consistently high scores (green). In the untruthful setting, the model introduces errors, resulting in sharp score drops (red).

    \item \textbf{Fairness}: We define fairness as unbiased response generation \citep{corbett2023measure}. Fig.~\ref{fig:fairness} shows an example with a man and a woman working together. The fair response is neutral, while the unfair one portrays gender bias. Token-wise scores drop at biased language.

    \item \textbf{Fearlessness}: Defined by confidence and reduced sensitivity to risk \citep{lilienfeld1996development}, fearlessness prompts the model to emphasize awe and beauty when describing an ocean scene (Fig.~\ref{fig:fearness}). Under fearful framing, language shifts toward danger and discomfort, revealing a conceptual inversion.

    \item \textbf{Power}: Associated with authority and dominance \citep{french1959bases}, power is examined through responses describing the U.S.\ Capitol Building (Fig.~\ref{fig:power}). A humble viewpoint emphasizes governance ideals, while a power-seeking perspective shifts toward control and manipulation.
\end{itemize}

\subsection{Attention Matrix Visualization}

\begin{figure}[tbp]
    \centering
    \includegraphics[width=0.5\textwidth]{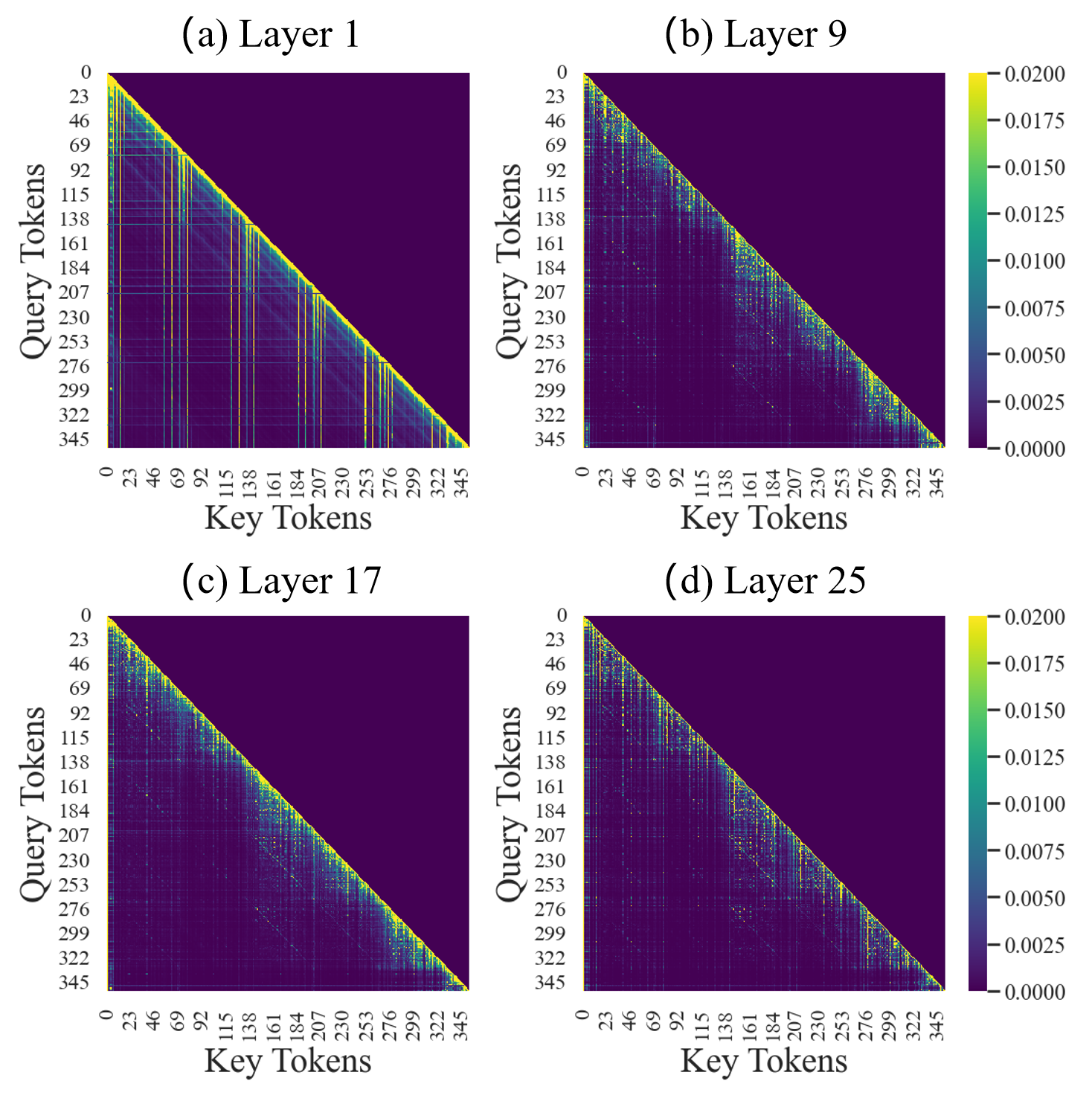}
    \caption{Attention matrix visualization across different layers.}
    \label{fig:pic1}
\end{figure}

Fig.~\ref{fig:pic1} visualizes attention matrices at various layers, illustrating that matrices become increasingly sparse in deeper layers. This sparsity likely arises as the model focuses on a smaller subset of crucial tokens, reducing the effective rank of the attention operator and clarifying the spectral structure.

\subsubsection{Evaluating Honesty and Fairness in VLMs}

To evaluate how well VLMs represent abstract ethical concepts, we analyze their handling of honesty and fairness in multimodal response generation, shown in Fig.~\ref{fig:combined_concepts}. To quantify this process, we compute token-wise projection scores following RepE \citep{zou2023representation}, measuring how closely activations align with concept directions at each layer.

Our method successfully identifies distinct conceptual behaviors within the model: when the VLM produces dishonest or unfair responses, token-wise projection scores show clear drops (red regions), in contrast to the consistently high scores observed for honest or fair cases. Such findings provide strong evidence that abstract ethical dimensions like honesty and fairness are internally structured and traceable.

\begin{figure*}[t]
    \centering
    \includegraphics[width=\textwidth]{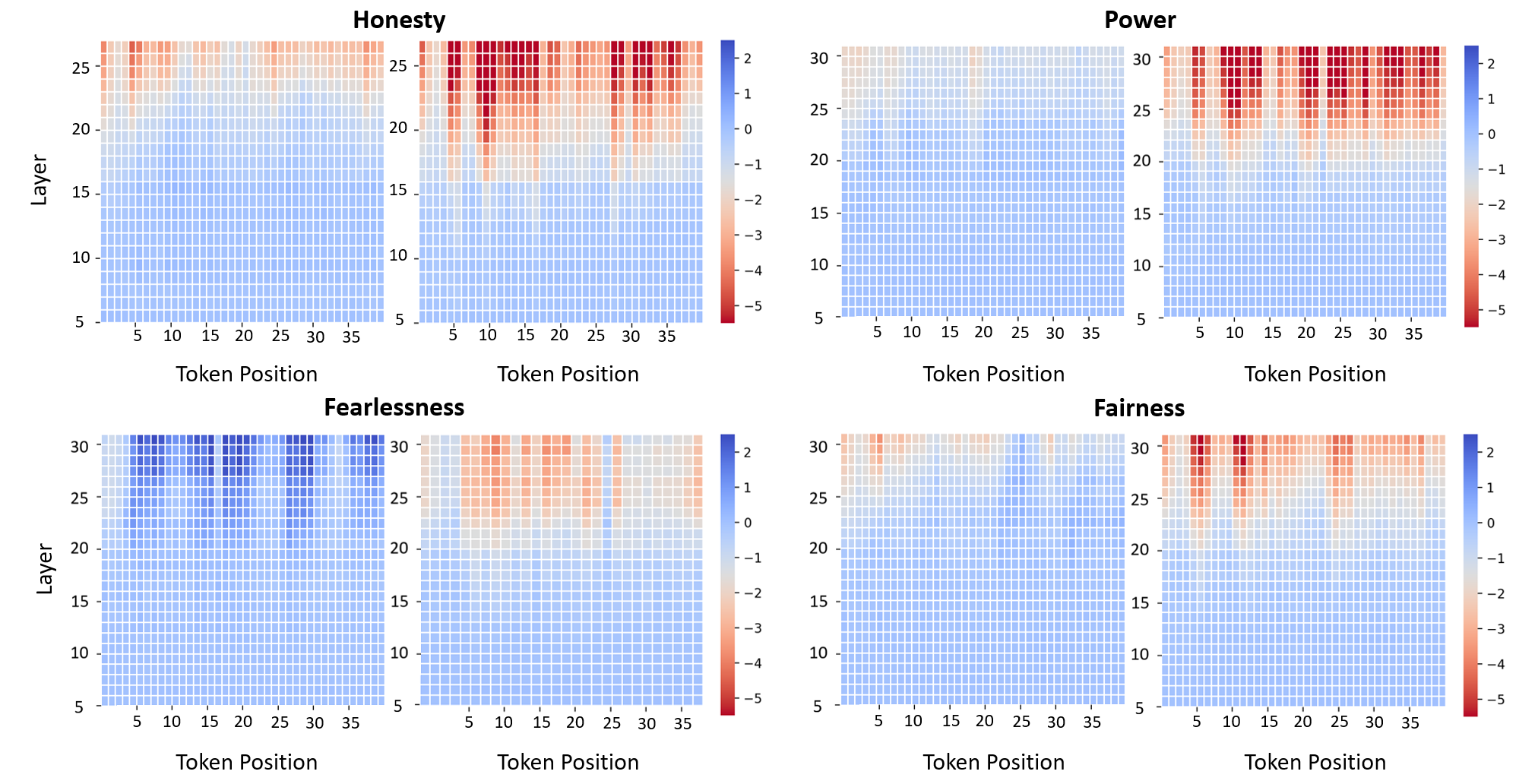}
    \caption{Temporal LAT Scans for Honesty, Power, Fearlessness, and Fairness. The left heatmap represents the LAT Scan when the VLM aligns with the concept, while the right heatmap corresponds to the opposing concept. Blue indicates high alignment, red represents low alignment.}
    \label{fig:LAT}
\end{figure*}

\subsubsection{LAT Scans for High-level Representations}

To capture how high-level concepts evolve and propagate through the model, we employ Linear Attribution Tomography (LAT) \citep{zou2023representation}, which enables layer-wise visualization of conceptual information flow. Fig.~\ref{fig:LAT} shows the resulting LAT scans.

The scans reveal concept-specific propagation patterns. Honesty and fairness exhibit stable trajectories under aligned prompts but greater dispersion and deviation under misaligned ones. Power appears concentrated in later layers, while fearlessness shows early-layer changes. These results are well explained by the SPP framework: concepts propagate through a small set of dominant spectral directions, and the stability observed in aligned cases reflects the spectral gap preservation predicted by \Cref{thm:spp_stability}. The consistency of these representations across modalities further demonstrates the robustness of RepE, and their traceability back to the input is explained by ISLH.

\end{document}